\newcommand{\sL}{\mathcal{L}}
\newcommand{\sM}{\mathcal{M}}
\newcommand{\sP}{\mathcal{P}}

\newcommand{\sU}{\mathcal{U}}
\newcommand{\sX}{\mathcal{X}}
\newcommand{\sY}{\mathcal{Y}}
\newcommand{\sR}{\mathcal{R}}

\newcommand{\R}{\mathbb{R}}

\newcommand{\is}{\vmathbb{1}_S}

\documentclass[manuscript, nonacm]{acmart}

\usepackage{multirow}
\usepackage{subcaption}

\AtBeginDocument{%
  }

\setcopyright{acmlicensed}
\copyrightyear{2026}
\acmYear{2026}
\acmDOI{XXXXXXX.XXXXXXX}
\begin{document}

%%
%% The "title" command has an optional parameter,
%% allowing the author to define a "short title" to be used in page headers.

% ORIGINAL:
% \title{Physics-Informed Blind Reconstruction of Dense Fields From Sparse Measurements Using Neural Networks With a Differentiable PDE Solver}

% SUGGESTED MODIFICATION:
\title{Leveraging Differentiable PDE Solvers for Semi-Neural Spatial Reconstruction From Sparse Measurements}

%%
%% The "author" command and its associated commands are used to define
%% the authors and their affiliations.
%% Of note is the shared affiliation of the first two authors, and the
%% "authornote" and "authornotemark" commands
%% used to denote shared contribution to the research.
\author{Ofek Aloni}
\authornotemark[1]
\email{ofek.aloni@campus.technion.ac.il}
\affiliation{%
  \institution{Department of Mathematics, Technion}
  \city{Haifa}
  \country{Israel}
}

\author{Barak Fishbain}
\affiliation{
  \institution{Department of Civil and Environmental Engineering, Technion}
  \city{Haifa}
  \country{Israel}
  }
\email{fishbain@technion.ac.il}

%%
%% By default, the full list of authors will be used in the page
%% headers. Often, this list is too long, and will overlap
%% other information printed in the page headers. This command allows
%% the author to define a more concise list
%% of authors' names for this purpose.
\renewcommand{\shortauthors}{Aloni \& Fishbain}

%%
%% The abstract is a short summary of the work to be presented in the
%% article.

\begin{abstract}
    Generating dense physical fields from sparse measurements is a fundamental question in sampling, signal processing, and many other applications. State-of-the-art approaches to this problem either rely on spatial statistics that ignore the governing physics, integrate the physics into a multiple-objective optimization process, or require examples of the complete, fully-resolved simulation state during training, which are frequently unavailable outside of synthetic benchmarks. Here, we present a novel alternative that leverages recent advances in the integration of numerical simulators with data-driven models. Namely, we propose a hybrid modeling pipeline that couples Radial Basis Function (RBF) reconstruction with a Neural Network (NN) correction and a Partial Differential Equation (PDE) solver, so that the numerical simulator itself is embedded directly in the training loop of the learned component. Notably, the NN is trained without assuming availability of examples of the fully-resolved simulation state. This is made possible by implementing the PDE solver so that it is end-to-end differentiable, allowing gradients to be backpropagated through the simulation step during training. This grey-box methodology is evaluated on three standard benchmarks from fluid mechanics, where it achieves superior results over statistical and machine-learning-based reconstruction methods.
\end{abstract}

\received{June 2026}
\received[revised]{}
\received[accepted]{}

%%
%% This command processes the author and affiliation and title
%% information and builds the first part of the formatted document.
\maketitle

\section{Introduction}

Inferring dense fields from sparse measurements is a fundamental phase in many scientific and engineering applications, as the number of sensors and spatiotemporal resolution of measurement is often limited. Examples of this task include the reconstruction of dense air pollution maps from distant measuring stations \cite{asaf-air-pollution} and remote sensing data \cite{zhangDeepSpatialPrediction2023}, soil moisture levels \cite{DeepSoil2026}, and surface elevation maps in ocean waves \cite{oceanWavesRecon2023}. This data scarcity motivates the use of post-hoc methods to approximate the full field of values from the available measurements.

Inverse problems of this nature, reconstruction of the dense field from sparse measurements, are inherently ill-posed, as multiple potential fields can correspond to the same sparse measurements. This ambiguity makes the domain a natural candidate for Machine Learning (ML) methods, which can learn to generate physically plausible solutions by capturing complex non-linear priors \citep{physmlreview2021}.

Classical reduced-order modeling techniques, such as Proper Orthogonal Decomposition (POD) and Dynamic Mode Decomposition (DMD), provide established frameworks for reconstructing state fields from sparse sensors \cite{schmidDynamicModeDecomposition2010}. These methods, particularly the Gappy POD approach \cite{eversonKarhunenLoeveProcedure1995}, operate by constructing a low-dimensional basis from a pre-computed library of full-field snapshots. The sparse observations are then projected onto this basis to resolve the ill-posed nature of the problem. These data-driven methods depend fundamentally on the availability of high-fidelity data.

Kriging spatial interpolation (or Gaussian Process Regression) is another classic statistical approach for dense field reconstruction \citep{SpatialStats}. Kriging provides the Best Linear Unbiased Predictor (BLUP) based on the spatial autocorrelation of the data, modeled via a variogram. While effective for stationary fields with well-behaved statistical properties, Kriging struggles with complex, non-linear changes in the sampled field.

Another prominent deterministic approach for spatial interpolation from scattered data is the use of Radial Basis Functions (RBFs) \cite{Buhmann_2003}. RBF methods reconstruct a continuous dense field by expressing it as a weighted sum of radially symmetric kernels centered at the sparse measurement locations. The weights are determined by solving a linear system that enforces exact interpolation at the sampled points.

An example of an ML approach for reconstruction is in Fukami et. al \cite{voronoicnn}. In their work, a Convolutional Neural Network (CNN) is employed to interpolate spatial data acquired in fluid dynamics problems, such as the vorticity field in a two-dimensional cylinder wake. The methodology proposed involves pre-processing the sampled data through the construction of Voronoi tessellations. This approach is fundamentally constrained, as a completely resolved field is typically unavailable to serve as ground truth for model training. Thus, this requires the use of simulated data for training purposes, wherein fully resolved fields can be sampled at the measurement locations. This requires a carefully calibrated model of the system, a non-trivial task in and of itself in real-world applications \cite{wangTowardDataCenter2023}. Although this method can produce fields that replicate similar physical behavior, the resulting initial conditions are not necessarily representative of those in the actual system. 

A prevailing methodology in physics-informed ML is to encourage the models to adhere to the physics of the problem by adding a loss term containing the residual of the Partial Differential Equation (PDE) governing the evolution of the field, as in Physics Informed Neural Networks (PINNs) \citep{raissi2017physics}. In PINNs, a NN is trained to approximate the spatiotemporal solution to a given PDE. Thanks to the PDE loss, the network is able to converge to a physically viable solution despite sparse spatiotemporal data. This comes at a cost of a more complex optimization setting. In addition, adherence to the PDE is not enforced; rather, it is encouraged via minimization of the loss term. Similarly, matching the sampled values at the sample location (exact interpolation) is not guaranteed.

To date, there have been limited efforts directly aimed at addressing the gap in ground truth data. For instance, Ehlers et al. \cite{ehlers-PINO-2025} investigated this challenge within the context of ocean waves by employing a Physics Informed Neural Operator (PINO) \citep{PINO}. In the study, a PINO is trained to reconstruct nonlinear wavefields in a single spatial dimension, utilizing only sparse buoy or radar measurements. Notably, the same buoy measurements are used as the input and for computing the data loss. This means the data is used only to encourage adherence, and the learning process is driven by the physics-based loss terms.

Connecting the domains of numerical simulation and optimization are \textit{differentiable PDE solvers} (physics simulators). They are end-to-end differentiable, meaning they can calculate gradients throughout the entire duration of a simulation for a given loss function with respect to desired parameters. This allows for integration with gradient-based optimization frameworks and deep learning pipelines. Example applications are in control and robotics \cite{heidenNeuralSim2021}, and fluid mechanics \cite{umSolverintheLoop2020}. In the broader sense, such an approach can be useful not only for PDE-based problems, but also in traffic simulation and optimization \cite{andelfingerTowardsDifferentiableAgentBased2023}. A growing number of dedicated physics simulation engines are available to implement differentiable solvers, such as DiffTaichi \cite{DiffTaichi2019} and PhiFlow \cite{phiflow}. For an extensive review of applications and available engines, see Newbury et al. \cite{newburyReviewDifferentiableSimulators2024}.

Leveraging the increasing accessibility of differential solvers, we introduce and evaluate a "semi-neural" pipeline for reconstruction of dense fields from sparse sensors. The sketch of the process is as follows:
\begin{enumerate}
    \item \textbf{Initial Reconstruction}: Radial Basis Functions (RBFs) are used to generate an initial guess for the dense field at time $t$.
    \item \textbf{Neural Network Correction}: A NN is applied to amend the RBF reconstruction. It is trained exclusively on the spatially sparse data (no full fields), using this exact pipeline in training. This is made possible thanks to the differentiability of the solver. 
    \item \textbf{PDE Solver}: A differentiable PDE solver is applied to the corrected reconstruction, thus introducing the physics, enforcing constraints, and evolving in time.
\end{enumerate}

This is what we call Sparse Physics-Informed REconstruction (SPIRE). The process has the following unique advantages:
\begin{itemize}
    \item \textbf{Grey-box model}: Numerical PDE solvers and the RBF reconstruction process are closed algorithms, that are well-researched and understood. This pipeline leverages the available data, while  limiting the role of a black-box neural network to correction.
    \item \textbf{Use of known physics}: Use of the simulator allows enforcing physical constraints. On their own, methods like RBF and Kriging do not offer ways to incorporate boundary conditions (BCs) and physical constraints. Even PINNs optimize for adherence to the PDE and the BCs, but do not directly enforce them. Use of the PDE solver offers a simple way to do so. We show that even without the NN, i.e. skipping step (2), use of the solver significantly reduces reconstruction error.
    \item \textbf{Train directly on measurement data}: The pipeline allows for training the NN directly on the measurement data, without relying on data generated externally.
    \item \textbf{No additional constraints on optimization}: Allows for usage of the sparse data for training without an additional PDE residual loss, simplifying the optimization landscape.
\end{itemize}

\section{Methodology}

\label{sec:methodology}

\subsection{Problem Formulation}
\label{sec:SPIRE-loop}
Let $\Omega \subseteq \mathbb{R}^d$ be the spatial domain of interest. The goal is to infer a spatiotemporal measurement field, given by an underlying function $u(t,x):\mathbb{R}^+\times\Omega\to \mathbb{R}^{n}$, that is a solution to some time-dependent partial differential equation of the form:
\begin{equation}\label{eq:pde-general}
    \frac{d}{dt}u+F[u]=0
\end{equation}

Where $F$ is a nonlinear differential operator. In the following, it is assumed that $F$ and the boundary conditions $\left.u\right|_{\partial\Omega}$ are known, but $u(t,x)$ is not. Instead, there is a set of $m$ fixed points in the domain, $ S\subset\Omega$ in which the value of $u$ is measured, while in the rest of $\Omega$ it is unknown. The sparse observations are denoted $s(t):=u(t,x)\cdot\vmathbb{1}_S$, where $\is$ is the indicator function:
\[
\is (x)=\begin{cases}
    1, \quad x\in S, \\
    0, \quad \text{otherwise}
\end{cases}
\]

Let $\mathcal{U}$ denote the space of spatial fields $f: \Omega \to \mathbb{R}^{n}$, representing possible states of the system at any fixed time. 

For the reconstruction tasks, we would like to deduce a reconstructing model, $\sM : \sU \to \mathcal{U}$ that, given a sparse measurement $s(t)$, returns an approximation of the full field:
\begin{equation}\label{ideal_M}
\sM\circ s(t) \approx u(t,\cdot)
\end{equation}

Note again that the problem of reconstruction is ill-defined. There could be two solutions $u,v\in \sU$ of the PDE $F$, such that $u\neq v$, but $u\cdot\is=v\cdot\is$. Thus, a recovery of the full field without additional assumptions or information is impossible. We can, however, attempt to construct $M$ such that $M\circ s\in\sU$ are physically consistent fields, while minimizing the error $\| \sM \circ s-u\|$.  

Working in a setup that mimics real-world measurement data, the sole input to the process is $s(t)$, with no samples of the entire field, $u(t,\cdot)$, known. Hence, implementing $M$ as a NN so $s(t)$ is the input and $u(t,\cdot)$ is the output to train against is not viable. The sample space $\sX=\{s(t_i)\}_{i=1}^N$ is available, but the target space $\sY=\{u(t_i,\cdot)\}_{i=1}^N$ is not. Thus, we cannot compute the data loss $\sL(M\circ s(t),u(t,\cdot))$ and are unable to learn an approximate reconstructor directly. 

This limitation is the main motivation to use time-dependence to train the model (section \ref{sec: training}). That is, given that the sensors measure the field over time, we use pairs of consecutive measurements: $s(t)$ as input, and $s(t+\Delta t)$ for supervision (computing the loss), where $\Delta t>0$ is small.

\subsection{Construction of the Reconstructor}

\subsubsection{Initial Reconstruction}

First, the sparse samples $s(t)$ are passed through a closed form reconstruction algorithm, denoted here as an operator $\sR:s(t)\mapsto u(t,\cdot)$. This serves both as an initial guess for the reconstructed field, and as a preprocessing step for the NN, inspired by the use of Voronoi tessellations in Fukami et al., \cite{voronoicnn}. Since this can be computed and stored before training the NN, computational speed is not a limiting factor. As such, we opted to use the best (lowest error) reconstruction algorithm available. As can be seen by the results in section \ref{sec:results}, this has proven to be Radial Basis Function (RBF) reconstruction \cite{Buhmann_2003}.

RBF methods reconstruct a continuous field by expressing it as a weighted sum of radially symmetric basis functions centered at the sample locations. Given sparse samples ${(x_i,u(x_i))}_{x_i\in S}$, the reconstructed field is written as
\[
[\sR\circ s(t)](x) = \sum_{x_i \in S} \alpha_i \, \phi\!\left(\lVert x - x_i \rVert_2\right)
\]
where $\phi(r)$ is a chosen radial kernel and the coefficients ${\alpha_i}$ are determined by enforcing interpolation at the sample locations. The weights are obtained by solving a linear system. In the present work, reconstruction with RBF is performed using the RBFInterpolator implementation in SciPy \cite{scipy}, with a thin plate spline kernel.

\subsubsection{Training Procedure} \label{sec: training}
\begin{figure}[t]
    \centerline{\includegraphics[width=8cm]{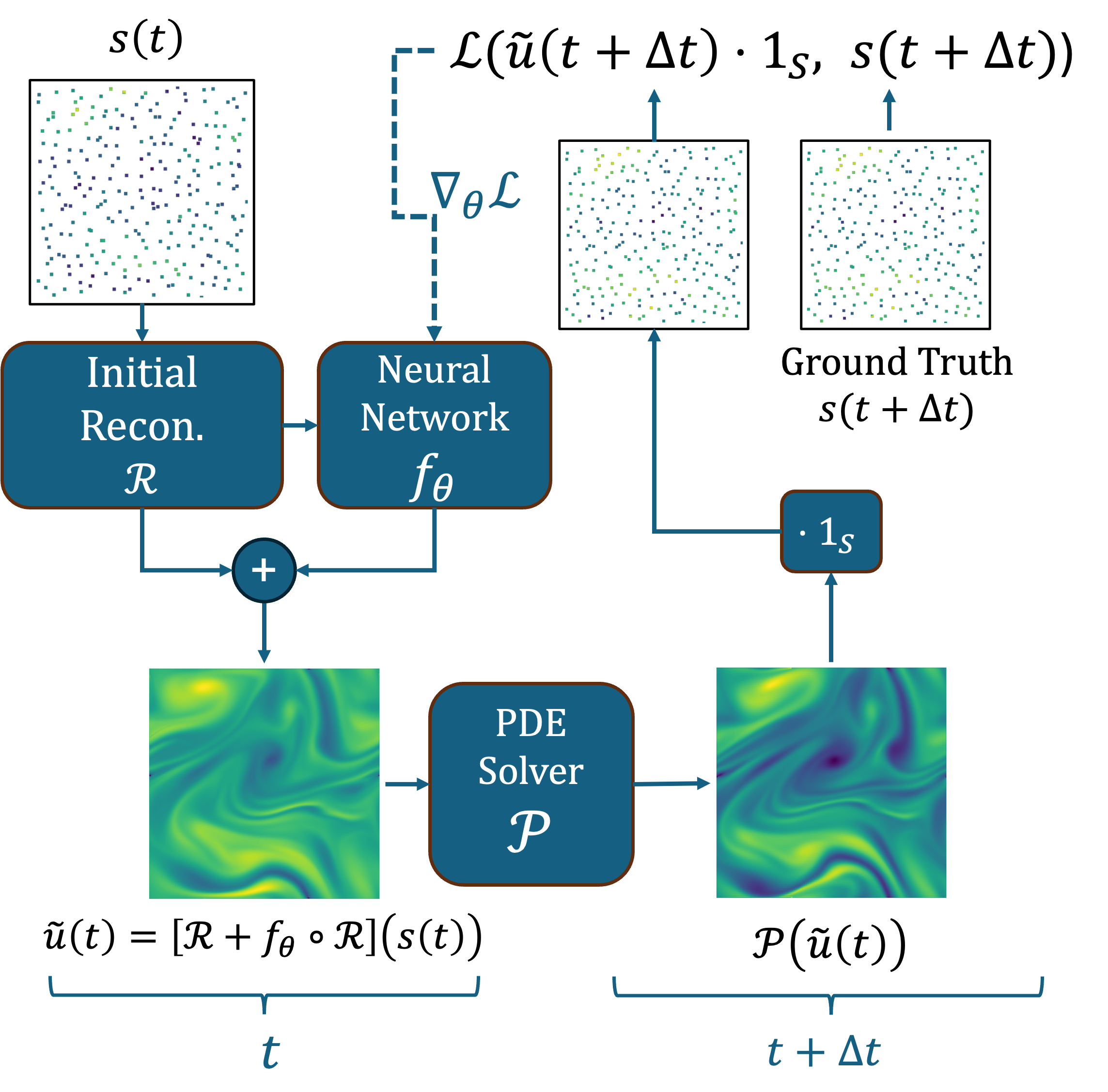}}
    \Description{A schematic of the SPIRE training loop. Sampled data is reconstructed, fed to the NN $f_\theta$, and run through the simulator. Then is sampled and compared to the ground truth at time $t+\Delta t$.}
    \caption{
    Schematic displaying the training process of $f_\theta$. First, an initial dense field is generated from the sparse samples $s(t)$ using RBFs, via the operator $\sR$. This is passed through the network $f_\theta$ and added as a skip connection. Then, it is advanced in time using the differentiable numerical solver $\sP$. Finally, the approximated field $\tilde{u}(t+\Delta t,\cdot)$ is sampled by multiplication with $\is$, and loss is compared relative to the ground truth field values at sample locations $s(t+\Delta t)$. 
    }
    \label{fig:SPIRE-schematic}
\end{figure}

Here, our objective is to train a neural network $f_\theta:\sU\to\sU$ as a residual operator that amends the initial reconstructions $\sR\circ s(t)$:
\begin{equation} \label{eq: corrected recon}
    \tilde{u}(t,\cdot)=[\sR+f_\theta\circ\sR](s(t))
\end{equation}

Using the NN for correction as in Eq. \ref{eq: corrected recon} can be framed as adding a skip connection between the input and output of the NN, inspired by ResNet \cite{heDeepResidualLearning2015} and subsequent architectures. 

Using the known PDE and BCs, $F,\left.u\right|_{\partial\Omega}$, a  numerical PDE solver $\sP:\sU\to\sU$ is implemented, for evolving the system in time:
\begin{equation}\label{solver}
    \sP(\tilde{u}(t,\cdot);F,\left.u\right|_{\partial\Omega}) =\tilde{u}(t+\Delta t,\cdot)
\end{equation}

Finally, the approximated, evolved field $\tilde{u}$ can be sampled in the sample locations via multiplication with $\is$. Then, the loss is computed relative to the true sampled values at time $t+\Delta t$:
\begin{equation} \label{eq:final-loss}
    \sL(\tilde{u}(t+\Delta t,\cdot)\cdot\is, s(t+\Delta t))
\end{equation}

For a visualization of the training loop for a single iteration on a single example, see Fig. \ref{fig:SPIRE-schematic}.

Computing the gradients $\nabla_\theta \sL$ that are needed to update the weights of $f_\theta$ during training relies on automatic differentiation \citep{autodiff}. In this respect, an important practical consideration is that $\sP$ must be differentiable, and compatible with libraries used for automatic differentiation (e.g. PyTorch \cite{pytorch}), to allow for backpropagation through the solver. There is an increasing number of solutions built to address this demand, such as PhiFlow \citep{phiflow}.

\subsection{Case Studies}
The performance of SPIRE is evaluated on three classic fluid mechanics problems: the 1D Burgers' Equation \citep{LinearNonlinearWaves}, 2D Wake flow (bounded Navier-Stokes with obstacle) \citep{WakeFlows1972}, and 2D Kolmogorov Flow (unbounded Navier-Stokes with periodic forcing) \citep{thess1992instabilities}.

\subsubsection{Burgers' Equation} \label{sec:burgers}
For a given field $u(x,t)$ and diffusion coefficient (or kinematic viscosity) $\nu$ , the general form of the viscous Burgers' equation in one spatial dimension is:
\begin{equation}
\partial_tu=\nu \partial_{xx}u -u \partial_xu
\end{equation}

In the simulated data, $\nu=0.01$, and periodic boundary conditions were used. The spatial domain was $\Omega=[0,1]$ with resolution $256$. $5,000$ initial conditions $u(t,x)$ were generated using Gaussian Random Fields (GRFs), as in \citep{fno}, and $u(t+\Delta t,x)$ was simulated with $\Delta t=0.01$. Diffusion and semi-lagrangian advection were computed using the PhiFlow library with PyTorch backend \cite{phiflow}. The number of samples was $|S|=32$, uniformly spaced in the interior of the domain.

\subsubsection{Wake Flow (Bounded Navier-Stokes with Obstacle)}\label{wake-flow-descrip}
In this setup, 2D incompressible flow around a fixed cylinder obstacle inside the domain is considered. We observe the velocities $\textbf{u}(x,y,t)=(u(x,y,t),v(x,y,t))$. The two-dimensional Navier-Stokes equations are of the form:
\begin{equation}\label{2d-ns}
    \partial_t\textbf{u}+\textbf{u}\cdot\nabla \textbf{u} =\nu\nabla^2\textbf{u} - \frac{1}{\rho}\nabla \textbf{p}
\end{equation}
With the additional constraint $\nabla\cdot\textbf{u}=0$ due to incompressibility. Note that the pressure $\textbf{p}$ is dependent on $\textbf{u}$. 

Data was generated by a numerical simulation with density $\rho=1$, and viscosity $\nu =10^{-5}$. A square $[0,1]\times[0,1]$ domain was discretized as $64\times 64$. The cylinder obstacle's radius was $R=0.075$, centered at $(0.5,0.7)$. Constant velocity on the upper boundary was set $v(y=1)= -0.02$, with a zero gradient boundary condition on the outflow $y=0$. Overall, the Reynolds number of such a setup, based on the cylinder diameter, is:
$$Re=\frac{v_\text{free}\cdot2R}{\nu}=\frac{0.02\cdot0.15}{10^{-5}}=300$$
In this range, periodic vortex shedding can be seen \citep{WakeFlows1972}. The simulated system ran for $500$ warmup steps. The data consists of 5,000 pairs with time difference $\Delta t=0.01$, with a step of $1$ between one initial condition and the next. Simulation using semi-lagrangian advection, explicit diffusion, and incompressibility enforcing was implemented using PhiFlow. The sensor sample locations $S$ were chosen uniformly, excluding points inside of the obstacle. The outside perimeter of the cylinder was also included in $S$. Overall, the number of sample locations was $|S|=247$. 

Note that using SPIRE, the condition $\mathbf{u}\equiv0$ inside the obstacle is inherently enforced via the solver $\sP$. There is no built-in way to enforce this in any of the other compared methods (see section \ref{baselines}). However, to evaluate the performance differences in non trivial parts of the domain, the velocity in the interior was set to zero post hoc, in all reconstructions. 

\subsubsection{Kolmogorov Flow (Unbounded Navier-Stokes with Periodic Forcing)}
Here, the problem observed is 2D incompressible flow with periodic boundary conditions, and periodic forcing in one of the axes:
\begin{equation}\label{forced-2d-ns}
    \partial_t\textbf{u}+\textbf{u}\cdot\nabla \textbf{u} =\nu\nabla^2\textbf{u} - \frac{1}{\rho}\nabla \textbf{p}+c\sin(ky)\hat{x}
\end{equation}

Where $c>0$ is a scale factor, and $k\in\mathbb{N}$ is the forcing wavenumber.  

In this setup, it is natural to observe the scalar vorticity field $\omega:=\nabla\times\textbf{u}$ rather than the velocities $\textbf{u}$ directly, since the velocities can be recovered from $\omega$, using stream functions. Hence, there exists a scalar $\psi(t,x,y)$ such that:
\[
u=\frac{\partial \psi}{\partial x}, v=-\frac{\partial \psi}{\partial y}
\]
Which means to recover $\textbf{u}$ from $\omega$, the following Laplace equation need to be solved:
\[
\omega=\nabla\times \textbf{u} =-\nabla^2\psi
\]
With the periodic boundary conditions and no obstacles the above is straightforward to solve numerically for $\psi$. As a consequence for the conducted tests, only the scalar field $\omega$ can to be reconstructed (in contrast to the vector velocity field in Wake flow). 

A $[0,1]\times[0,1]$ domain with resolution of $256\times256$ was chosen. Data was generated using $\rho=1$, $\nu=10^{-3}$, forcing with scale factor $c=100$ and wavenumber $k=8$, using time increment $\Delta t=0.001$. The solver was implemented directly in PyTorch, using a spectral approach based on the algorithm in \cite{koehler2024}. After $500$ warmup steps, $5,000$ consecutive samples were saved. Sample locations are uniformly spread in the domain, with $|S|=3,200$.

\subsection{Comparison Methods}\label{baselines}
We compare our model with existing methods for reconstruction. Due to the scarcity of ML-based reconstruction methods that are trainable using the assumed data availability, enabling a fair comparison, the method is also compared to non-ML methods for reconstruction. There, reconstruction from the sparse samples is based on spatial correlations in the sampled fields, and in particular does not require any training examples.

\subsubsection{Physics-Informed Neural Networks (PINNs)} \label{sec:pinns}
As a primary comparison method, we consider Physics-Informed Neural Networks (PINNs) \cite{pinn2019}. The core idea of PINNs is to train a neural network to approximate the spatiotemporal solution to a given PDE, guiding the model to a physically viable solution that matches the available data by incorporating a PDE residual loss. For a given PDE operator $F[u]$ as in eq. \ref{eq:pde-general}, the added loss term is:
\begin{equation}
    \sL_{\text{PDE}} = \sum^{N_r}_{i=1} |\frac{d}{dt}f_\theta(t_i,x_i) + F[f_\theta](t_i,x_i)|^2
\end{equation}

An advantage of comparing to PINNs is the relative wealth of existing research on the method, including common pitfalls and workarounds. We follow best practices outlined in Wang et al. \cite{wangExpertsGuideTraining2023}. This includes random weight factorization, Fourier embeddings, self-adapted loss weighting, and splitting the temporal domain to respect causality. Furthermore, the same model architecture and hyperparameter choices are used for Kolmogorov and Wake flows. 

We adapt the handling of the input data to match the problem formulation. Here, data-driven loss was computed in each iteration on a batch of randomly selected subset of measurements $\{u(x_i,t_i)\}_{i=1}^{N_{batch}}$, where $x_i\in S$ and $t$ were randomly sampled per batch. This is in contrast to using samples only from the initial and boundary conditions. In Burgers, the data is not taken from a continuous simulation (see section \ref{sec:burgers}), and so a PINN is individually trained for each initial-final state pair. In Wake and Kolmogorov, the temporal domain was divided into 10 and 5 separate time windows, respectively, as in \cite{wangExpertsGuideTraining2023}, and different, architecturally identical PINNs were trained on each time window. This was done to reduce the difficulty of optimization.

We note that the SPIRE architecture (Section \ref{sec: training}) and the PINN architecture were not subjected to identical hyperparameter search budgets: the PINN configuration follows the best-practice recommendations of \cite{wangExpertsGuideTraining2023} directly, whereas $f_\theta$'s architecture and initialization were tuned on the specific case studies considered here. We believe the gap in Table \ref{tab:results} is unlikely to be fully attributable to tuning effort alone, given that it is consistent with documented PINN training pathologies on multi-scale and vortex-shedding flows reported independently of this work \cite{chuangExperienceReportPhysicsinformed2022a}, but we flag the asymmetry for transparency.

\subsubsection{Inverse Distance Weighting (IDW)} IDW is a deterministic method that approximates values at unsampled locations by a distance-weighted average of the available samples. For an arbitrary point $x \in \Omega$, the reconstructed value $\hat{u}(x)$ is given by:
\begin{equation}
    \hat{u}(x) = \frac{\sum_{x_i \in S} w_i u(x_i)}{\sum_{x_i \in S} w_i}, \quad w_i = \frac{1}{\|x - x_i\|_2^p}
\end{equation}
where $p>0$ is a power parameter controlling the rate of decay. Note that taking the limit of $p\to\infty$ is equivalent to assigning each point with the value of the nearest neighbor (i.e. computing the Voronoi tessellation). In Burgers and Kolmogorov, $p=3$, and in Wake flow, $p=4$, selected to minimize $L^2$ error.

\subsubsection{Kriging} Next is Kriging, also known as Gaussian Process Regression \cite{SpatialStats}. It relies on the spatial autocorrelation of the data, treating the field $u$ as a realization of a stochastic process composed of a deterministic trend $\mu(x)$ and a stochastic residual $\varepsilon(x)$. The method provides the Best Linear Unbiased Predictor (BLUP) by solving a system of linear equations based on a modeled semi-variogram. A typical distinction is between Ordinary Kriging, which assumes the underlying trend $\mu(x)$ is constant but unknown (stationary mean), and Universal Kriging, which relaxes this assumption by modeling $\mu(x)$ as a polynomial function of the spatial coordinates. The latter allows the model to account for systematic spatial drifts that may be present in the data. Another important choice is of the functional form of the variogram model, e.g. gaussian, linear, exponential, and more. 

Interpolation and fitting of the variogram is done using PyKrige \cite{pykrige}. For wake flow, universal kriging with a spherical variogram model is used, while for Kolmogorov, Ordinary Kriging with exponential variogram. In Burgers, a 1d Ordinary Kriging with spherical variogram is used. In every instance, the variogram model and Kriging type were selected to minimize $L^2$ error.

\subsection{Implementation Details}
PyTorch \cite{pytorch} is used to carry out the experiments. For the 2D cases, $f_\theta$ is an Attention U-net CNN \cite{attention-Unet}, a five-level encoder–decoder with feature widths of 64–1024 channels with kernel size 3. At each decoder stage, the skip connection from the corresponding encoder level is filtered by an additive attention gate before concatenation. For the 1D case, a simple CNN is trained, with 7 layers, each with 48 filters, and kernel size of 16. In all cases, the loss function is the Mean Square Error (MSE), and training is conducted with the Adam optimizer \citep{zhang2018improved} with a fixed learning rate. Learning rate and batch size were, respectively, $10^{-3}, 100$ in Burgers, $10^{-4}, 64$ in Wake, and $0.25, 32$ in Kolmogorov. Of the $5,000$ realizations generated per case study, 4,000, 500, and 500 samples were used for training, validation, and testing respectively. Training was run for up to 100 epochs, with an early stop criterion based on validation error.

To mitigate exploding gradients and enhance the stability of training in the 2d cases, weights were initialized with uniform Xavier initialization \cite{xavier_init} with low gain, as in \cite{thuereyWelcomePhysicsbasedDeep}. Gain was set to $0.1$ (Kolmogorov) or $10^{-5}$ (Wake), as well as initializing biases to zero. This is particularly well suited to this application, as initializing the network to be close to the trivial transformation $f_\theta \equiv 0$ reduces to simply applying RBF to the sampled data. This gives the PDE solver a reasonable initial guess (see section \ref{sec:results}), which is important for the numerical stability of both the solver and the training process.

All training and testing procedures were conducted on a local Windows-based workstation equipped with an Intel i7 9700 8-core CPU and 96 GB of RAM, and a CUDA-enabled NVIDIA GeForce RTX 2060 GPU with 8 GB of memory.

Unless otherwise noted, all results in Table \ref{tab:results} correspond to a single training run of $f_\theta$ per case study, with variability reported across the 500 test samples rather than across independent training runs.

\section{Results}\label{sec:results}
\subsection{Evaluation of Reconstruction}
\begin{table*}[ht]
  \caption{Mean relative $L^2$ error of the reconstruction $\pm$ standard deviation, calculated on an a test set containing 500 samples, in each of the compared methods - the SPIRE pipeline (ours), trained PINNs, Kriging, IDW, and, RBF. Lowest error in bold. In parentheses - an ablation, of using the reconstruction methods for initial reconstruction, and applying the simulator without a NN. As can be seen, SPIRE outperforms the compared methods across all test cases.}
    \label{tab:results}

  \begin{center}
    \begin{small}
      \begin{sc}
        \begin{tabular}{lccccc}
        \toprule
        & \textbf{SPIRE} (Ours)         & \textbf{PINN}   & \textbf{RBF}                 & \textbf{Kriging}    & \textbf{IDW}                                                \\ \hline
        Burgers                 & $\mathbf{0.017} \pm 0.006$ & $0.05 \pm 0.03$ & $0.024 \pm 0.011$            & $0.2 \pm 0.2$     & $0.09 \pm 0.02$                                             \\
        &                            &                 & ($0.019 \pm 0.006$)           & ($0.2 \pm 0.2$)     & ($0.04 \pm 0.01$) \\ 
        Wake Flow               & $\textbf{0.043} \pm 0.003$            & $0.27\pm0.01$   & $0.064 \pm 0.003$            & $0.078 \pm 0.003$   & $0.108 \pm 0.002$                                           \\
        &                            &                 & ($0.053 \pm 0.003$) & ($0.066 \pm 0.004$) & ($0.093 \pm 0.003$)                                         \\
        Kolmogorov  & $\mathbf{0.10} \pm 0.02$   & $2.1 \pm 0.8$   & $0.12 \pm 0.02$              & $0.3 \pm 0.2$       & $0.2 \pm 0.02$                                              \\
        &                            &                 & ($0.11 \pm 0.002$)           & ($0.3 \pm 0.2$)     & ($0.2 \pm 0.02$) \\                          

        \bottomrule
        \end{tabular}      
        \end{sc}
    \end{small}
  \end{center}
  \vskip -0.1in
\end{table*}

The performance of each reconstruction model across the various test cases is summarized in Table \ref{tab:results}. For each scenario, $5000$ realizations are generated under varying initial conditions. These realizations are subsequently partitioned into training, validation, and test sets, the latter containing 500 samples $\{s(t_i)\}^{500}_{i=1}$.  The reconstruction performance of SPIRE and all comparative methods is assessed via the relative error on the test set with respect to the ground truth fields; specifically, for a reconstruction $\tilde{u}$ of $u$, the scalar $\frac{\|\hat{u}-u\|_2}{\|u\|_2}$ is evaluated. Reported are the mean relative error $\pm$ standard deviation across all reconstructions.

SPIRE consistently achieved the lowest mean relative errors in all evaluation scenarios, with RBF reconstruction (without any additional processing) as next-best. This motivates the decision to use RBF for initial reconstruction.

Additional analysis was done to confirm that SPIRE's improvement over the next-best baseline is statistically significant, and not driven by a small number of outlier samples. Pairwise comparison of the relative error of SPIRE compared to each of the baselines revealed that SPIRE has the lower error on almost all samples. In Wake, SPIRE outperforms every single baseline on every single test sample. In Kolmogorov, RBF had the lower error on only three test samples, and in Kriging on a single sample. Burgers' RBF baseline was the most competitive, with lower error on 50 out of the 500 test samples. Still, a one-sided Wilcoxon signed-rank test applied to the errors in this comparison confirms that SPIRE's reduction in error is statistically significant ($W=3870, p=4\times 10^{-74}$).

As an ablation testing the effect of the NN amendments, a similar pipeline that does not include a NN was tested. This means the initial field was reconstructed, and then ran through the solver, $\sP \circ\sR(s(t))$. The result of this ablation can be seen in Tab. \ref{tab:results}, in parentheses. In all cases, applying the solver had at worst no effect on the error. In most, the solver reduced the reconstruction error, in one case (Burgers with IDW) by as much as $56 \%$ .

\begin{figure*}[htbp]
\centering
\setlength{\tabcolsep}{6pt} % Horizontal padding between columns
\renewcommand{\arraystretch}{1.0} % Brought down from 1.2 to reduce default padding

\begin{tabular}{ccc}
    \small \textbf{Ground Truth} & \small \textbf{SPIRE (Ours)} & \small \textbf{PINN} \\[-0.8em]
    \raisebox{-\height}{\includegraphics[width=0.31\textwidth]{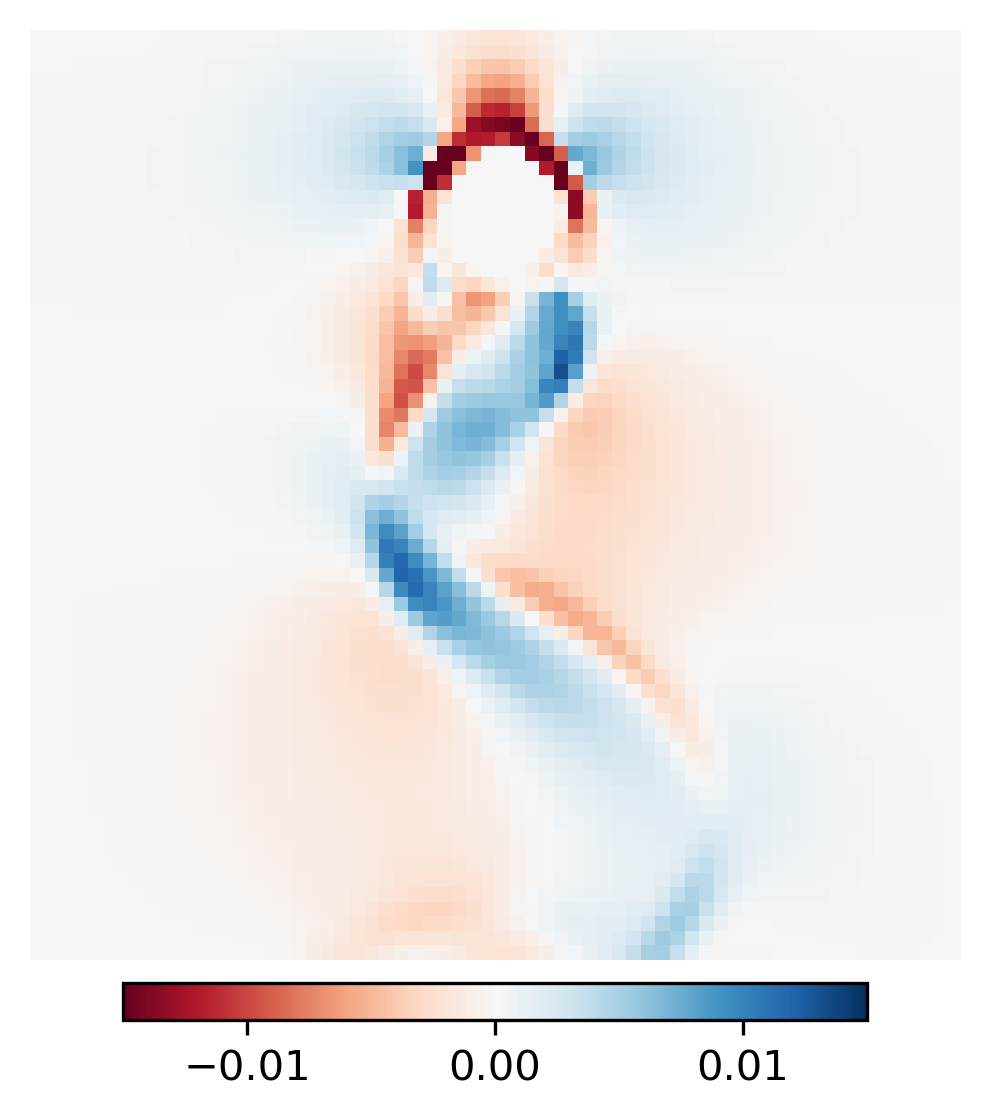}} &
    \raisebox{-\height}{\includegraphics[width=0.31\textwidth]{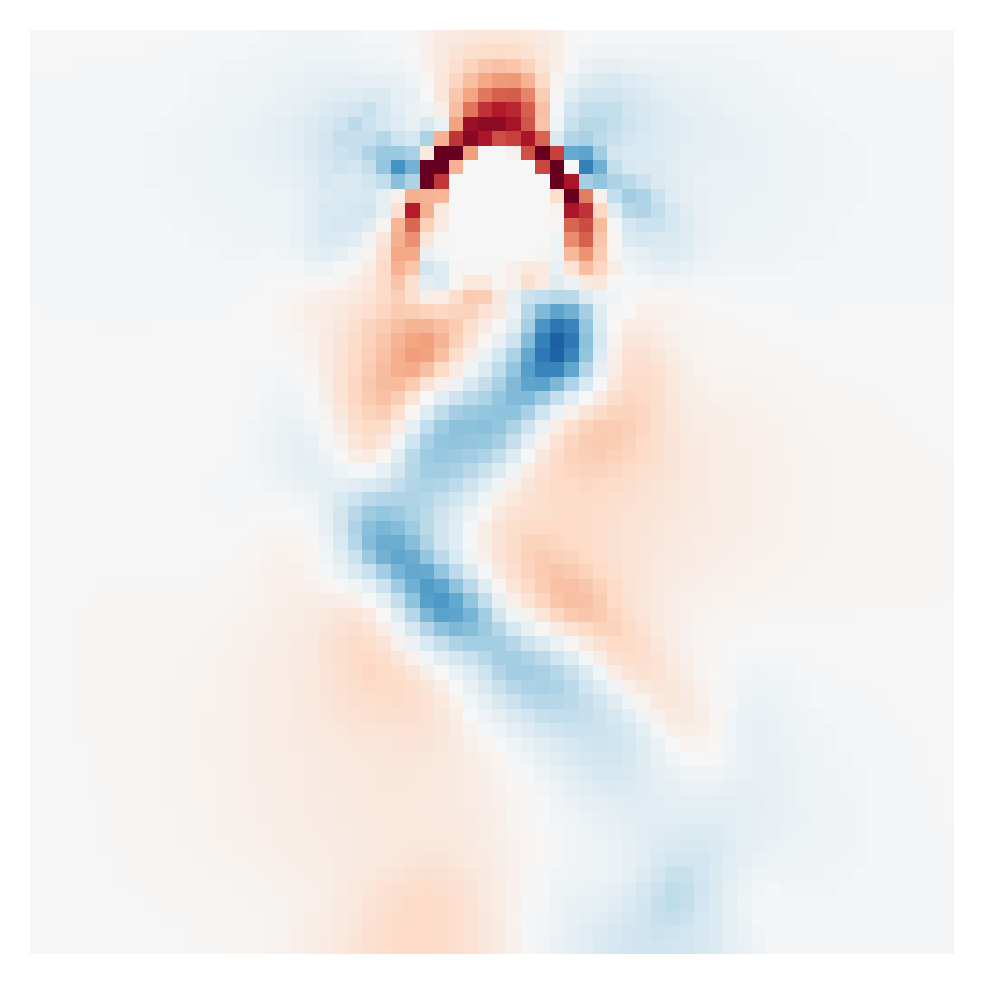}} &
    \raisebox{-\height}{\includegraphics[width=0.31\textwidth]{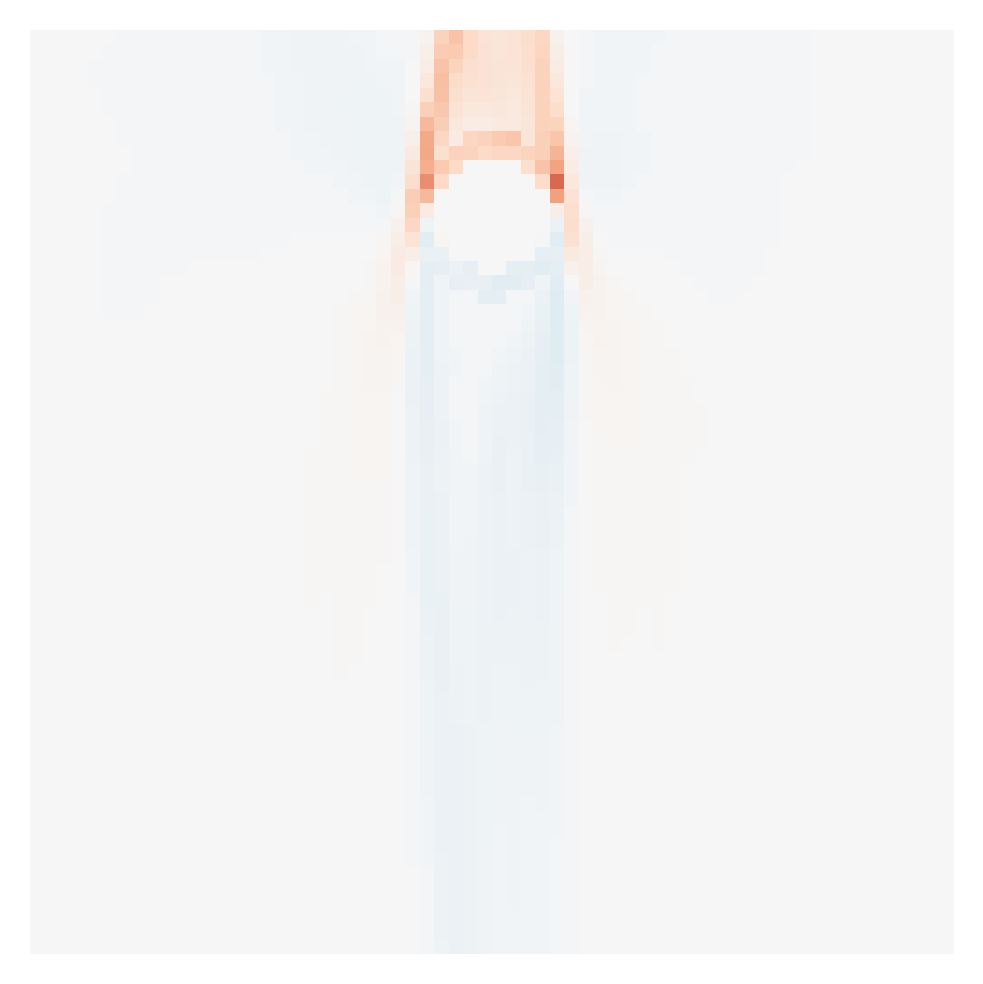}} \\

    \\[0.05em]
    
    \small \textbf{RBF} & \small \textbf{Kriging} & \small \textbf{IDW} \\[-0.8em]
    \raisebox{-\height}{\includegraphics[width=0.31\textwidth]{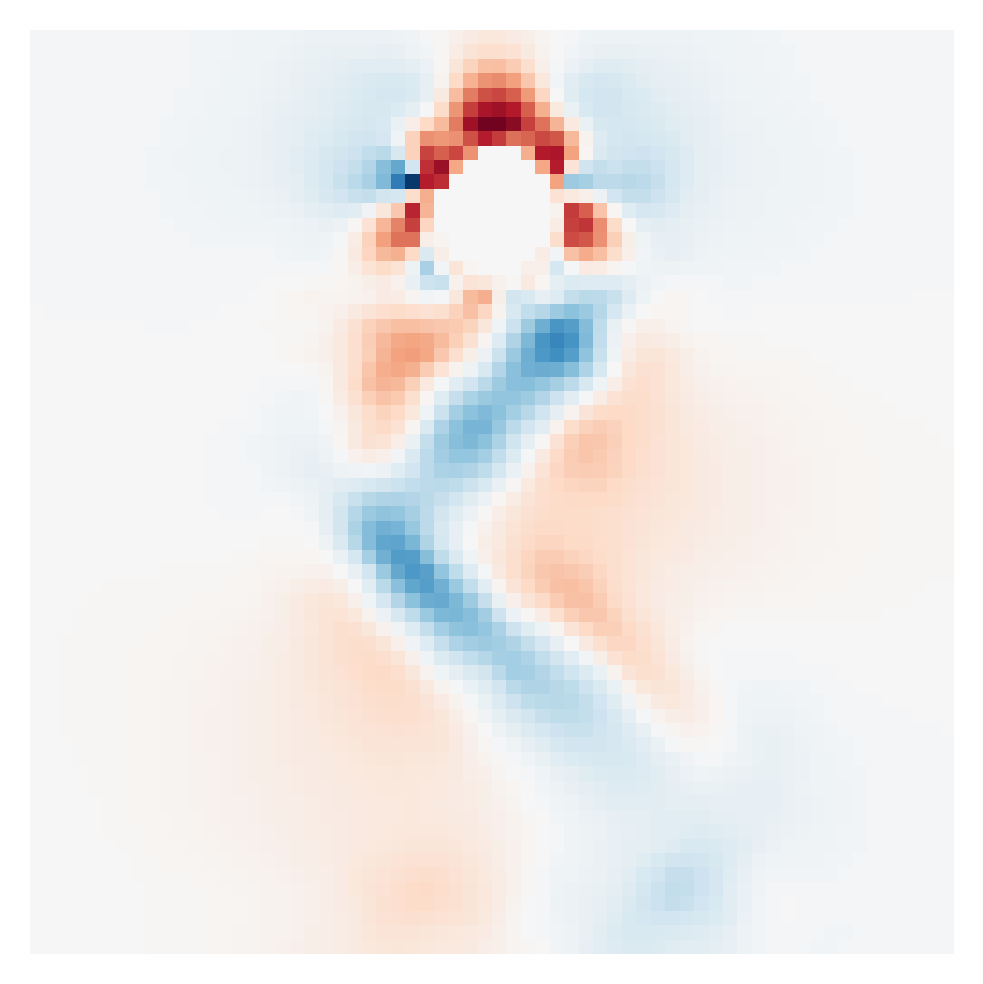}} &
    \raisebox{-\height}{\includegraphics[width=0.31\textwidth]{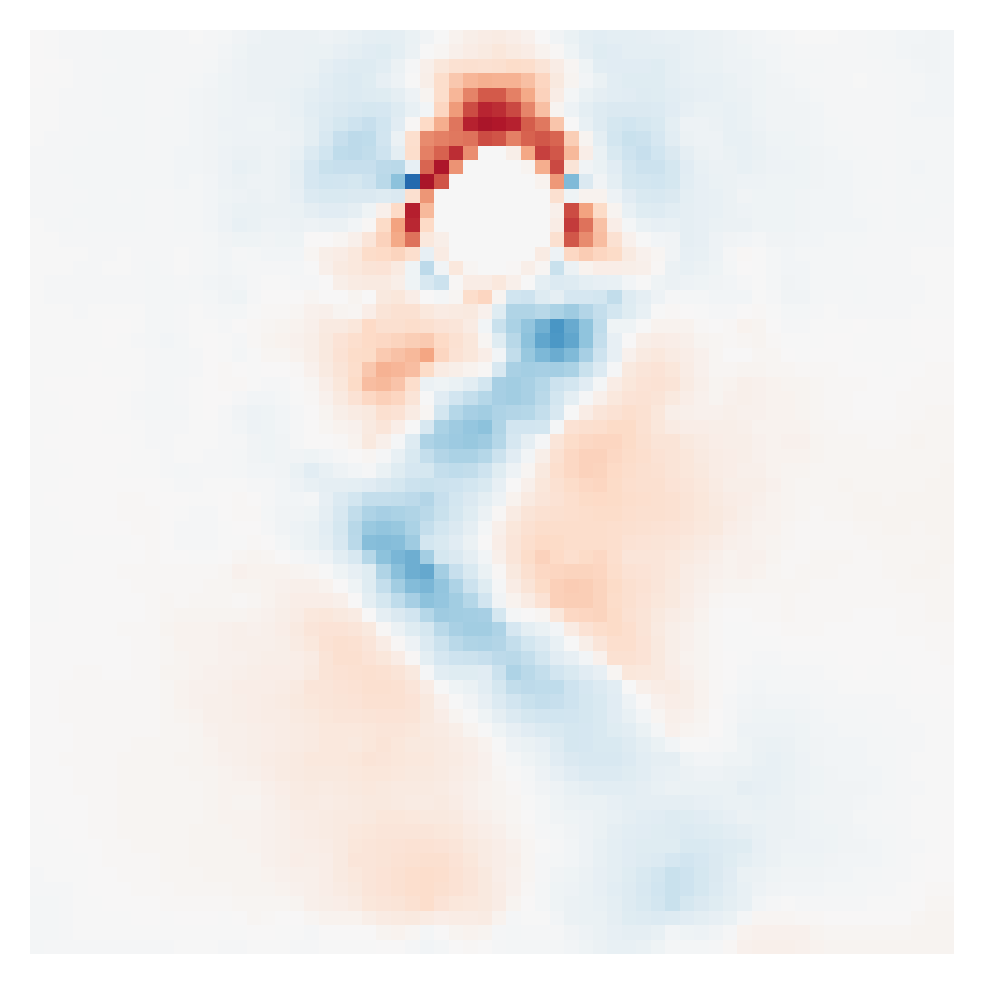}} &
    \raisebox{-\height}{\includegraphics[width=0.31\textwidth]{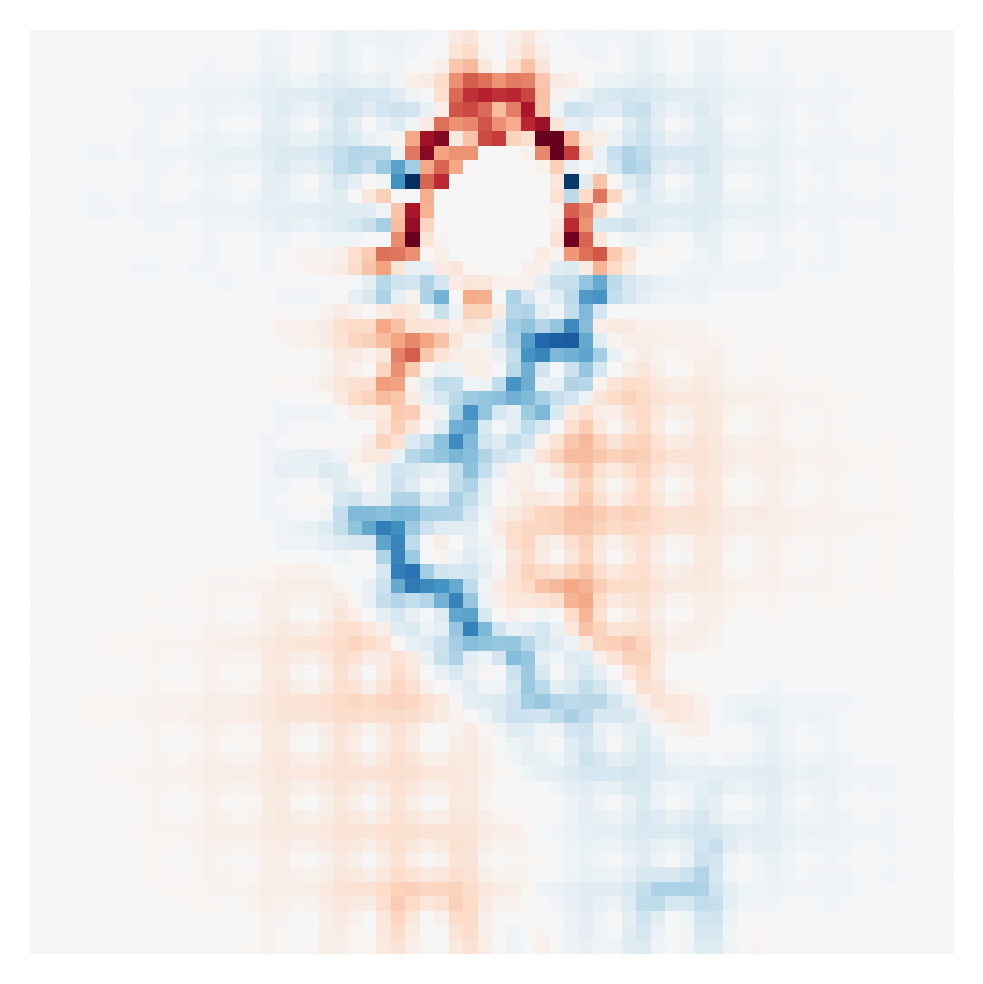}} \\
\end{tabular}
\Description{A comparison of the vorticity of reconstructions of a single test sample in Wake, showing the ground truth compared to SPIRE and the baselines. SPIRE is the most visually precise in reconstruction around the obstacle. RBF and Kriging are comparable, while PINN fails to qualitatively show vortex shedding, predicting laminar flow instead.}

\caption{Wake flow (Incompressible flow with obstacle) - visual comparison between the vorticity $\omega=\nabla\times\mathbf{u}$ of reconstructions of $\mathbf{u}(t+\Delta t,x,y)$. Top row (left to right): ground truth, SPIRE and PINN. Bottom row: RBF, Kriging, and IDW. Vorticity is shown for visualization only. Heatmap range is fixed across all subplots according to the ground truth.}
\label{fig:wake-results}
\end{figure*}

The performance disparity in the Wake scenario is especially pronounced. PINNs suffer from the highest error, as they fail to qualitatively capture the behavior of the system (Fig. \ref{fig:wake-results}). SPIRE offers a decrease of $49 \%$ in average error compared to directly applying RBF reconstruction. In addition, recall that errors in Table \ref{tab:results} are computed after the zero velocity condition in RBF is manually imposed (see \ref{wake-flow-descrip}). In the absence of this constraint, the velocity field inside the cylinder is non-zero, resulting in a mean relative error of $0.09$, more than doubling the error compared to SPIRE.

\begin{figure}[ht]
  \begin{center}
    \centerline{\includegraphics[width=0.8\linewidth]{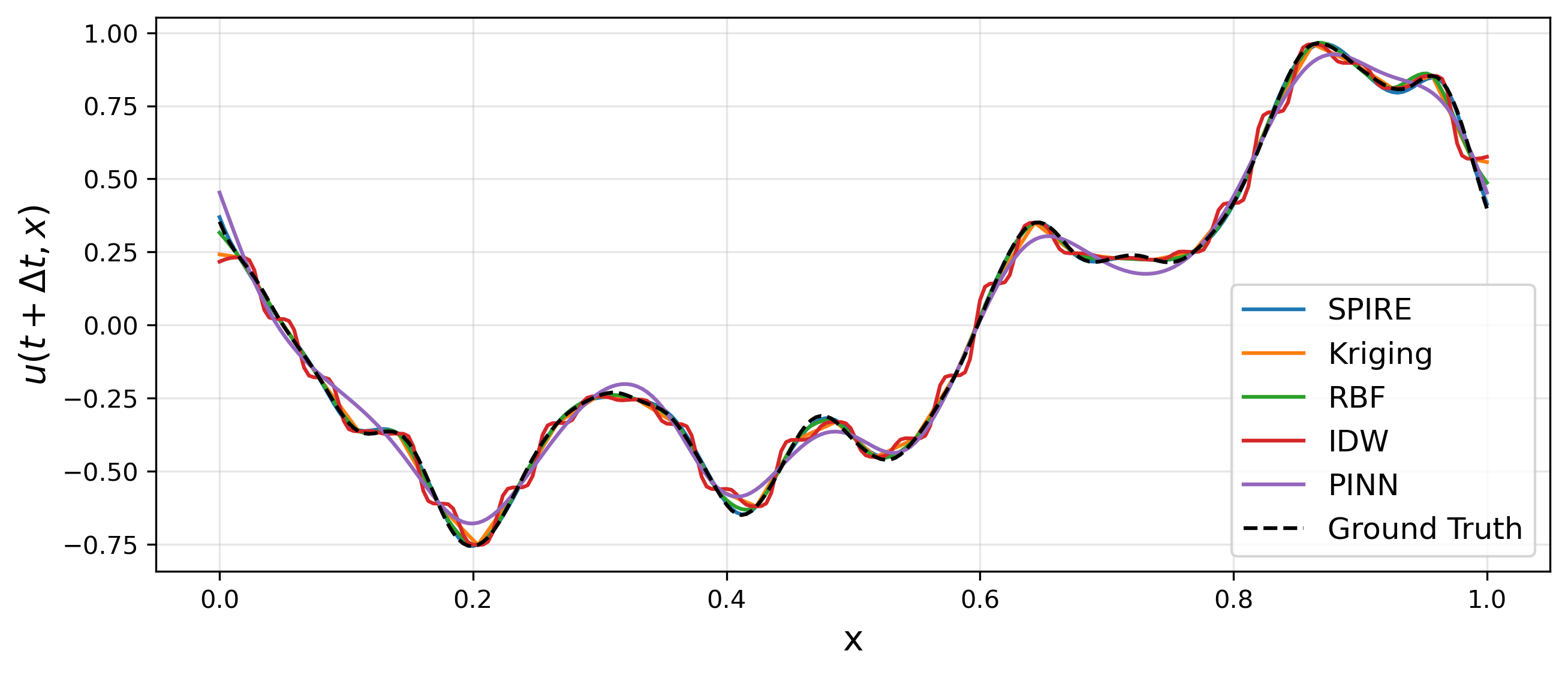}}
    \Description{A comparison plot, showing the field $u(t+\Delta t)$ vs $x$, and particularly a single example of the ground truth and the reconstructions of the different methods. SPIRE is the most accurate, especially around the periodic BCs.}
    \caption{1d Burgers equation - reconstruction of a single sample $s(t)$ from the test set by the each method, compared to the ground truth $u(t+\Delta t,x)$.
        }
    \label{fig:burgers}
  \end{center}
\end{figure}

% WAKE - GRID LAYOUT

For the Burgers equation, Fig. \ref{fig:burgers} illustrates the reconstruction of a single sample $s(t+\Delta t)$ from an initial sample $s(t)$ by each method. SPIRE, RBF, and Kriging closely follow the ground truth across most of the domain. Notably, near the boundaries $x=0,1$, SPIRE aligns more closely with the true field than the other approaches. IDW, Kriging, and RBF do not provide periodic fields. The PINN reconstruction, in contrast, does adhere to the periodic BCs, but strays the furthest from the true field (e.g. around $x=0,0.2,0.4$). 

\begin{figure}[ht]
    \centering
    % First subfigure
      \begin{subfigure}[b]{0.48\columnwidth}
        \centering
        \includegraphics[width=0.55\linewidth]{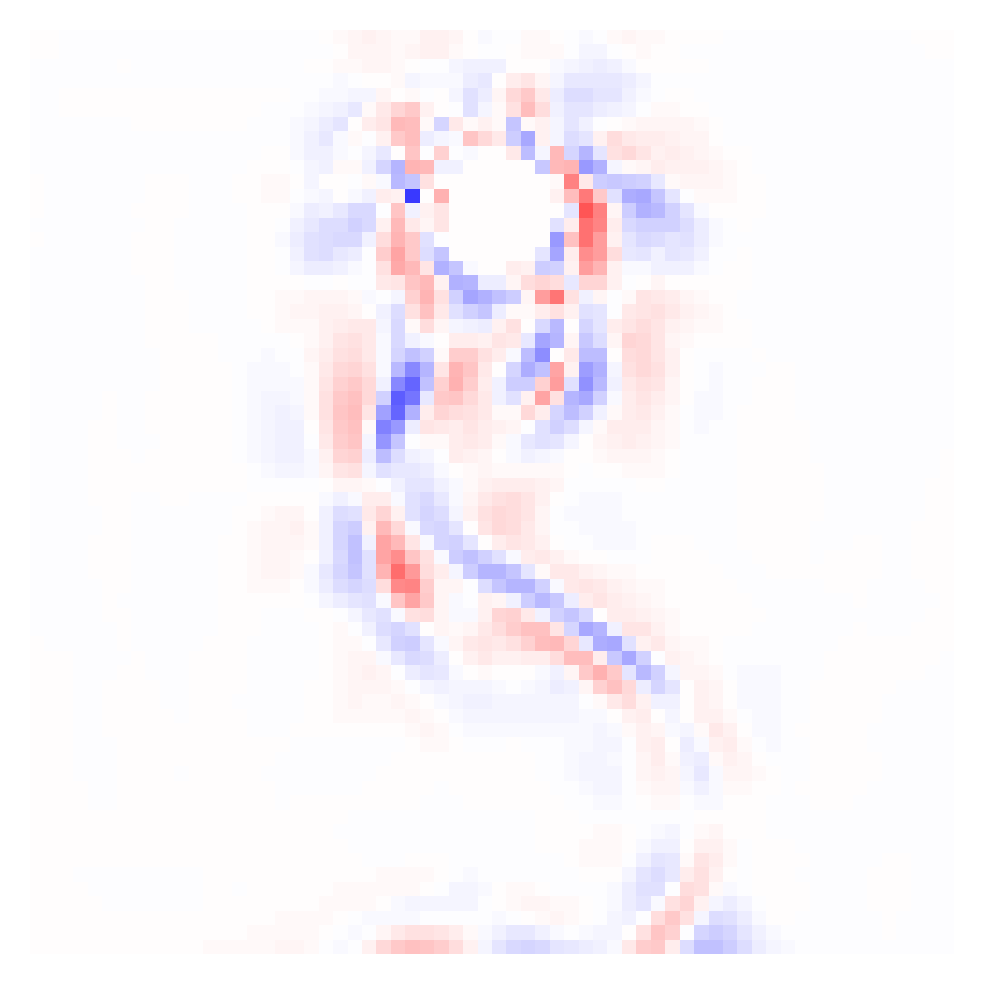}
        \caption{SPIRE}
        \label{fig:SPIRE-residual}
    \end{subfigure}
    \hspace{-1.0in}
      % Second subfigure
      \begin{subfigure}[b]{0.48\columnwidth}
        \centering
        \includegraphics[width=0.7\linewidth]{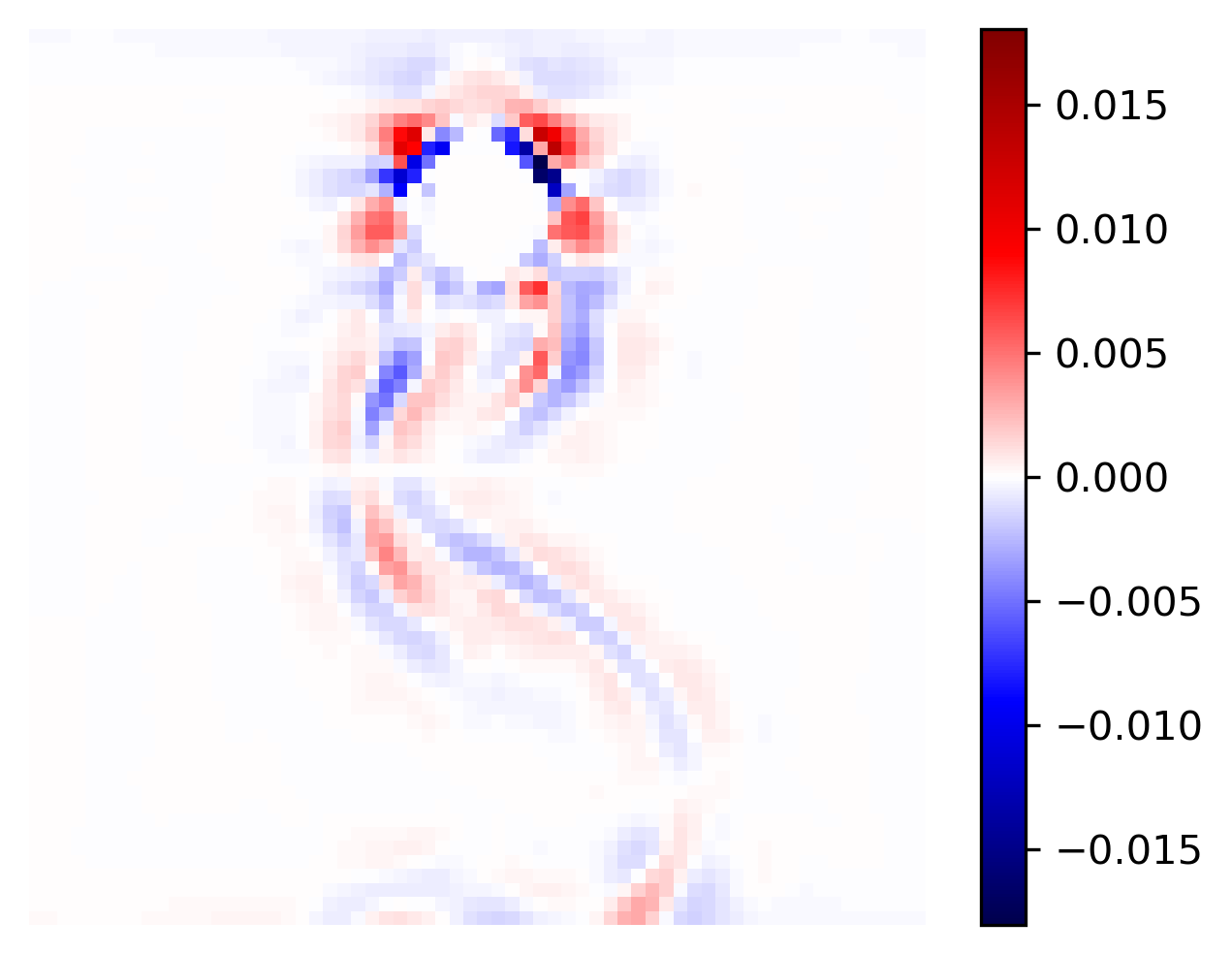}
        \caption{RBF}
        \label{fig:rbf-residual}
      \end{subfigure}
    \Description{A plot of the residual error on a single example from Wake. It emphasizes how SPIRE reduces the error around the obstacle.}
    \caption{Pointwise residual error between the true field and the reconstruction, for both RBF reconstruction (right) and SPIRE (left) in Wake flow. Note the lower error around the cylinder in SPIRE.}
    \label{fig:residual-comparison}
\end{figure}

To provide a detailed visualization of the reconstructions in the Wake flow scenario, Fig. \ref{fig:wake-results} presents the vorticity $\omega = \nabla \times \mathbf{u}$ associated with a single velocity field $\mathbf{u}(t+\Delta t, x, y)$, as reconstructed by SPIRE and the comparison methods. Unlike the Kolmogorov case, where $\omega$ is the directly measured and reconstructed variable, here the velocity field is reconstructed, and vorticity is displayed for visualization purposes only. PINNs fail to to capture the vortex shedding phenomenon, and reconstruct a laminar flow. We remark that this happens despite our best efforts at PINN implementation, and such results has been previously reported in the literature \cite{chuangExperienceReportPhysicsinformed2022a}. The improvement in SPIRE compared to RBF is most prominent around the cylinder. Note the "ears" to the left and right, and the shape of the flow above. This is emphasized in Fig. \ref{fig:residual-comparison}, showing the residual $\omega - \tilde{\omega}$ between the true field and the reconstruction of both RBF and SPIRE. 

% KOLMOGOROV FIGURE
\begin{figure*}[hb]
\centering
\setlength{\tabcolsep}{6pt} % Horizontal padding between columns
\renewcommand{\arraystretch}{1.0} % Brought down from 1.2 to reduce default padding

\begin{tabular}{ccc}
    \small \textbf{Ground Truth} & \small \textbf{SPIRE (Ours)} & \small \textbf{PINN} \\[-0.8em]
    \raisebox{-\height}{\includegraphics[width=0.31\textwidth]{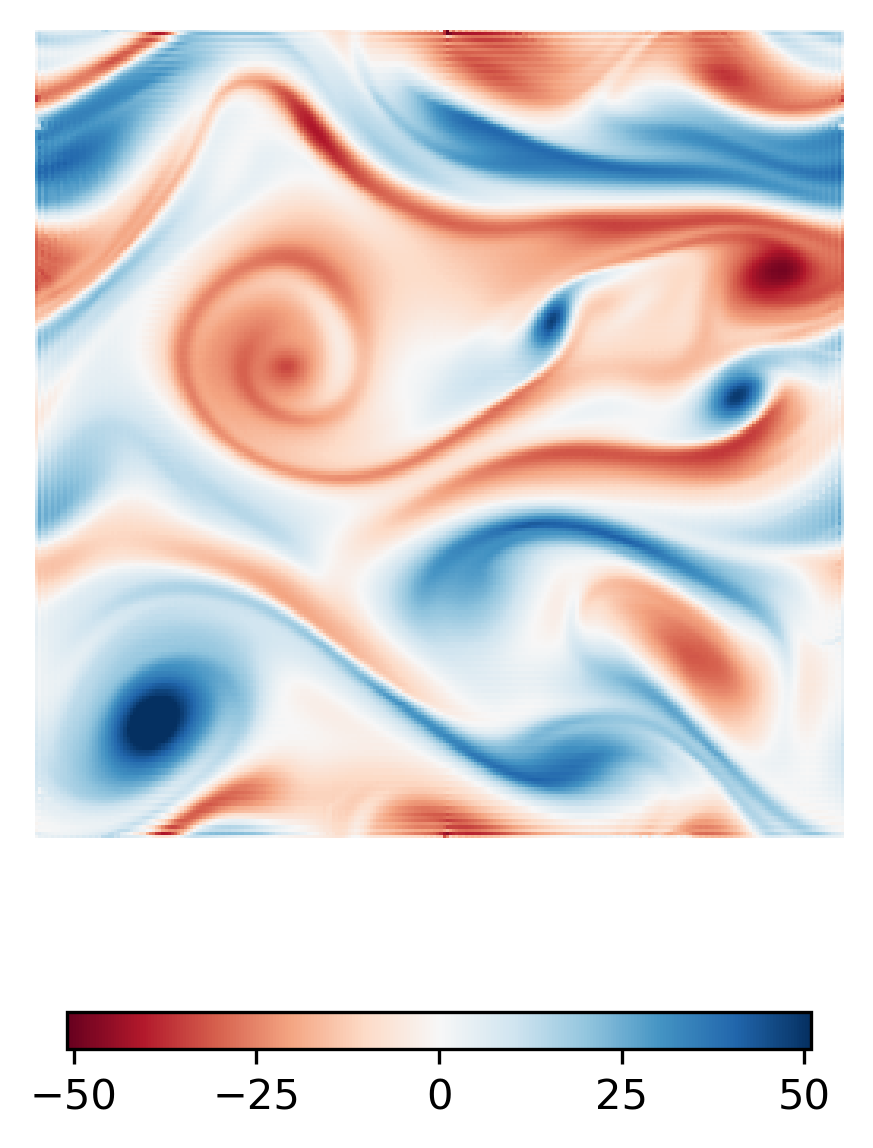}} &
    \raisebox{-\height}{\includegraphics[width=0.31\textwidth]{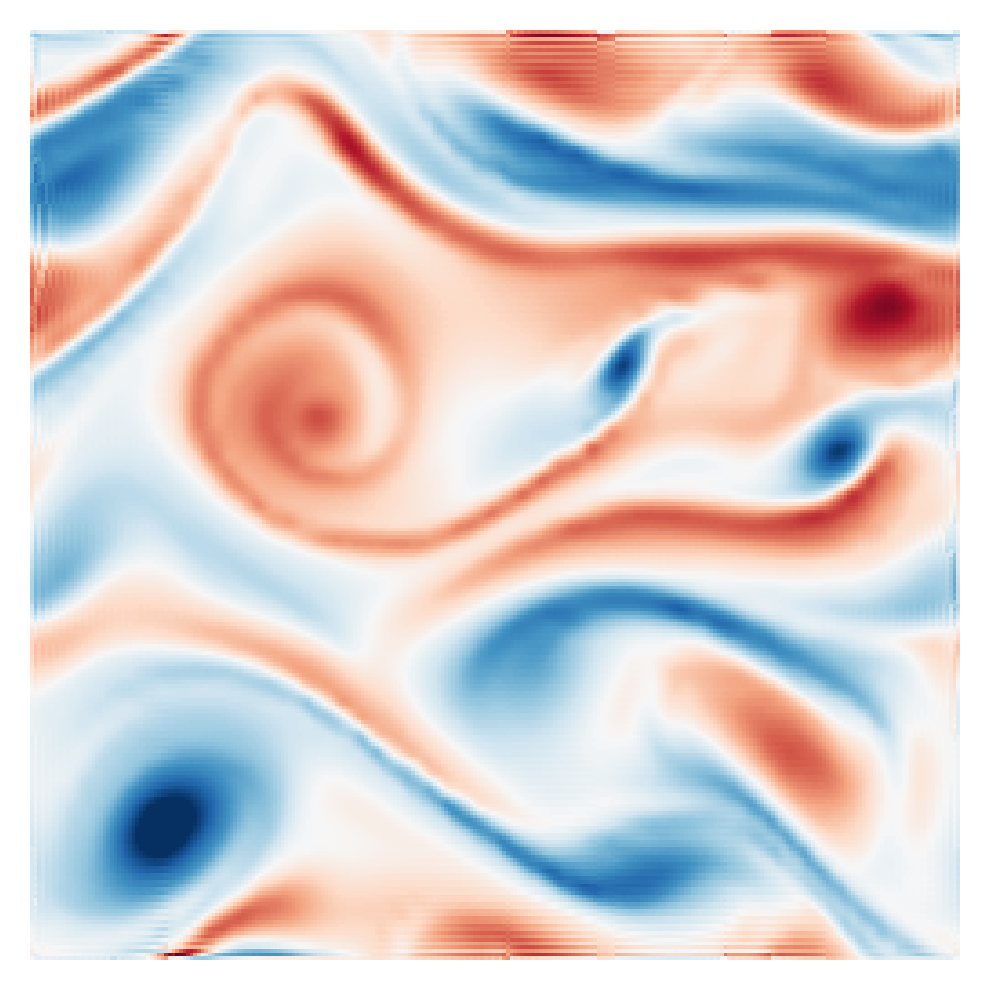}} &
    \raisebox{-\height}{\includegraphics[width=0.31\textwidth]{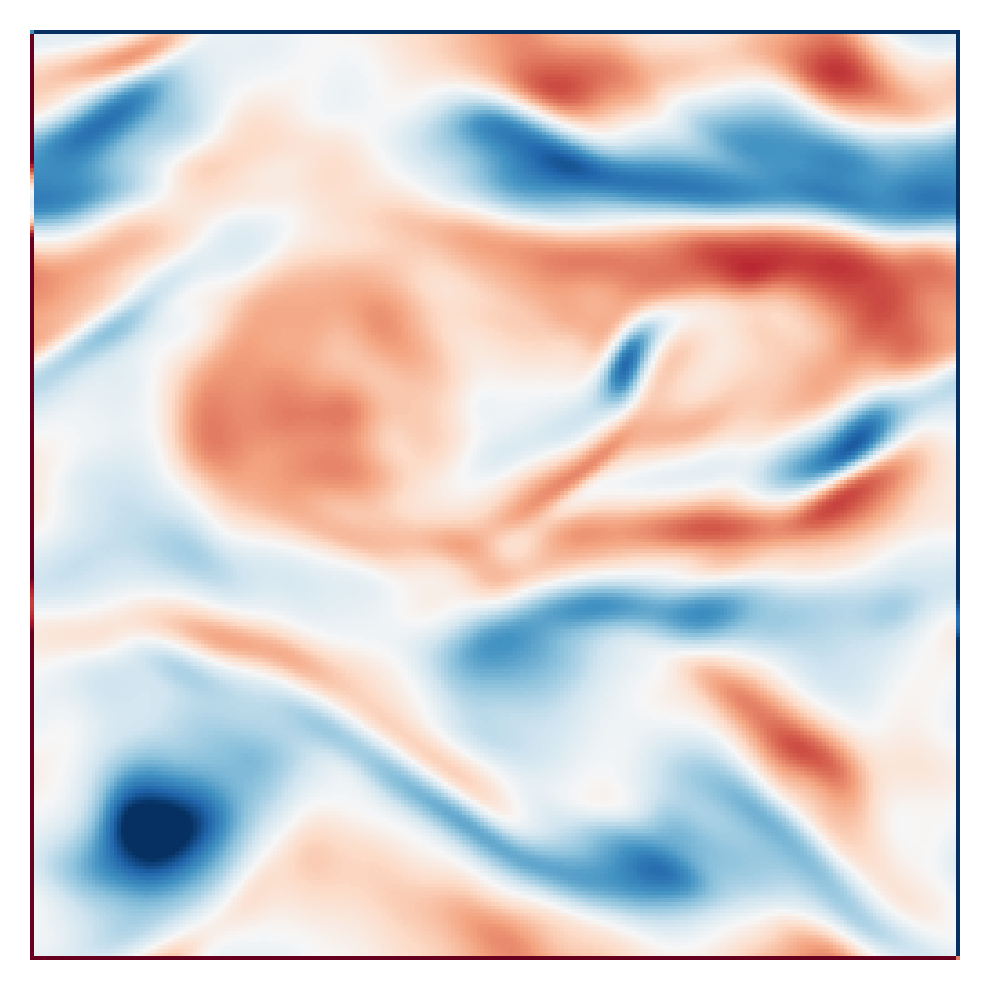}} \\

    \\[0.05em]
    
    \small \textbf{RBF} & \small \textbf{Kriging} & \small \textbf{IDW} \\[-0.8em]
    \raisebox{-\height}{\includegraphics[width=0.31\textwidth]{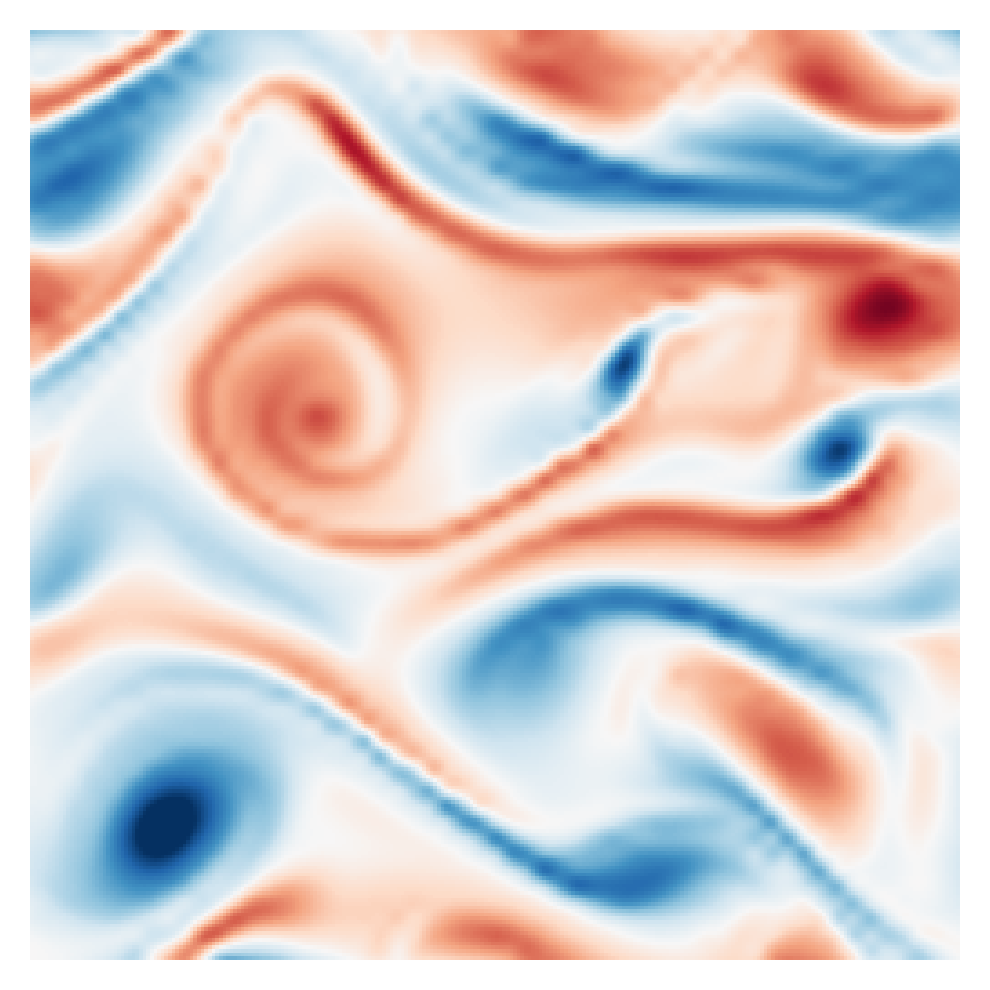}} &
    \raisebox{-\height}{\includegraphics[width=0.31\textwidth]{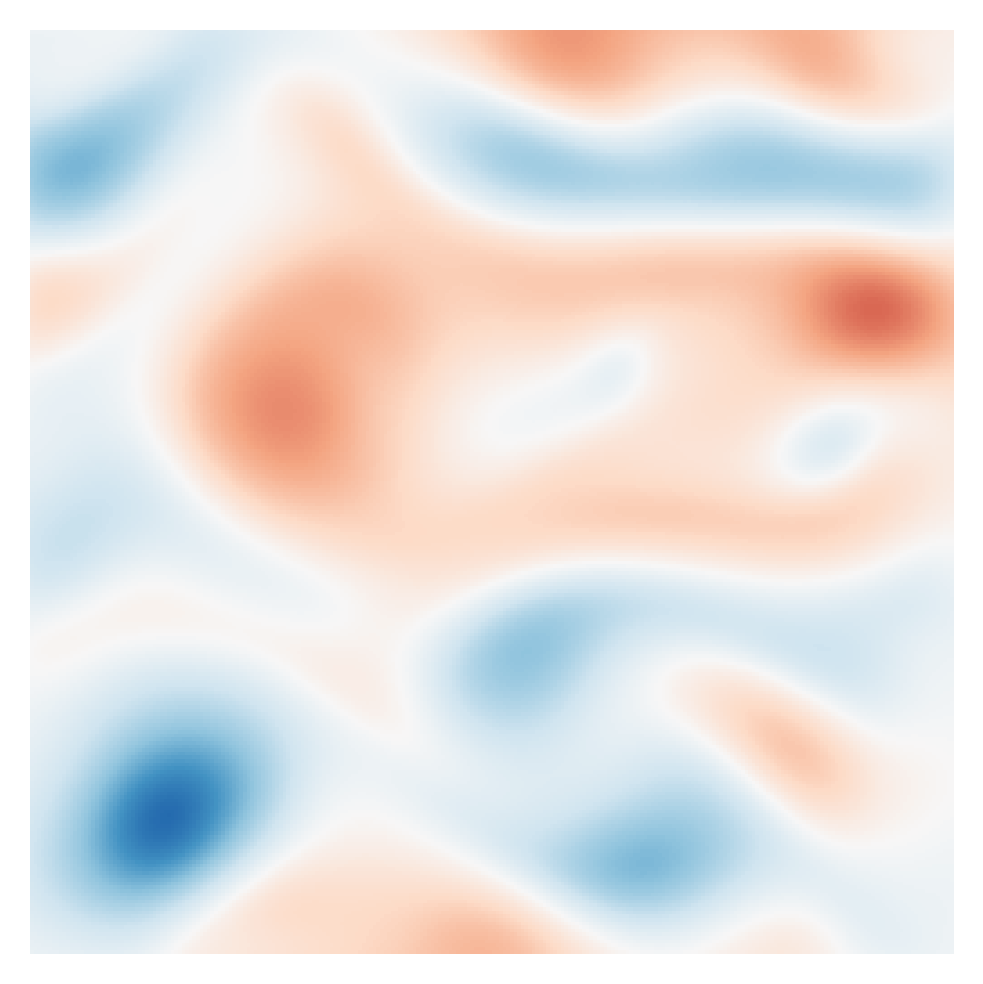}} &
    \raisebox{-\height}{\includegraphics[width=0.31\textwidth]{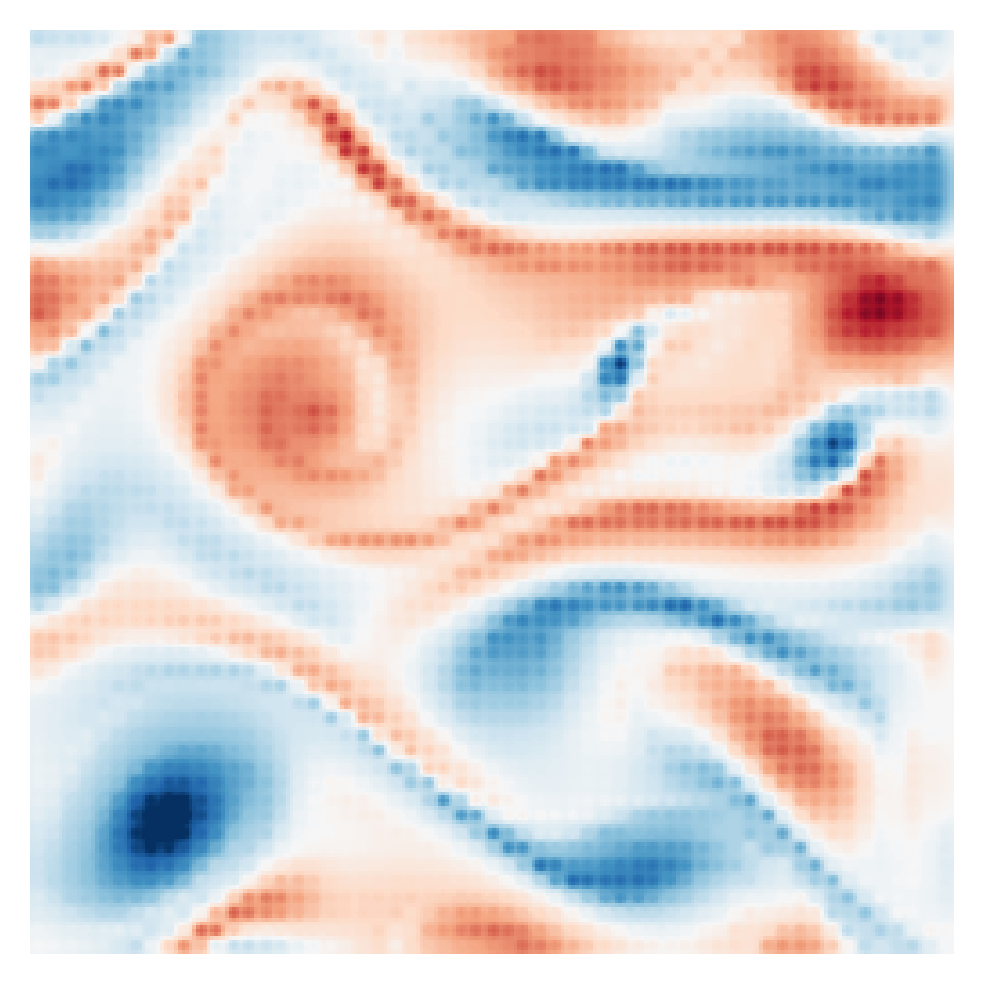}} \\
\end{tabular}

\Description{Comparison of the reconstructed vorticity field in Kolmogorov. SPIRE and RBF are comparable, but SPIRE is more accurate on edges in the vorticity field.}

\caption{Visual comparison of reconstruction of the vorticity field $\omega = \nabla \times \mathbf{u}$ in Kolmogorov flow. Top row (left to right): ground truth, SPIRE and PINN. Bottom row: RBF, Kriging, and IDW. Heatmap range is fixed across all subplots according to the ground truth.}
\label{fig:kolmogorov-full-comparison}
\end{figure*}

\begin{figure}[hb]
    \centering
    \includegraphics[width=0.3\linewidth]{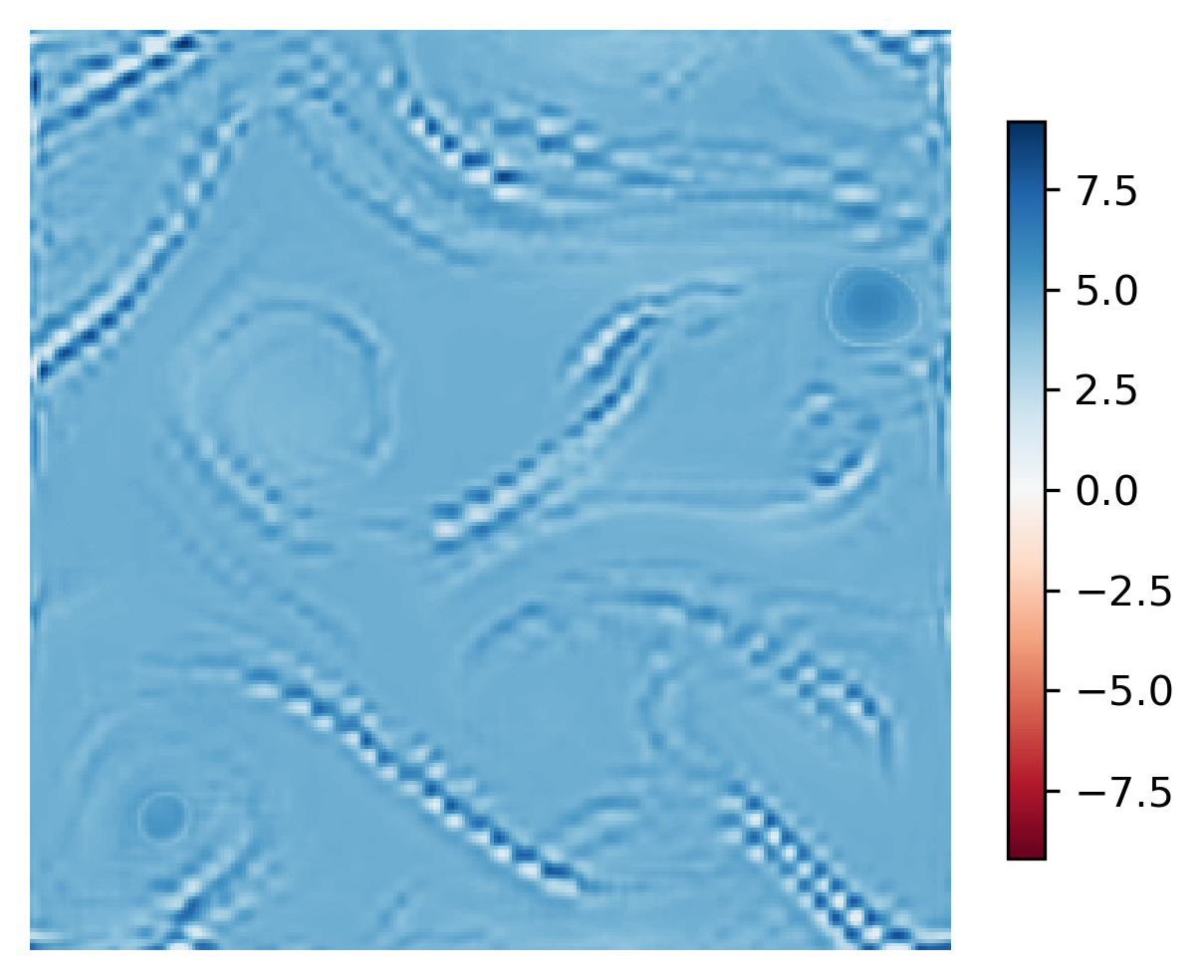}
    \Description{Plot with the raw NN output of the Kolmogorov test sample. It shows the main contribution is in the peaks of the field, and on edges in the vorticity, correcting staircase patterns in the RBF reconstruction.}
    \caption{Raw NN output for the vorticity field in Kolmogorov, used in the pipeline for the SPIRE reconstruction (same field as in Fig. \ref{fig:kolmogorov-full-comparison}). Note how the network learns to correct the jagged output given by the RBF.}
    \label{fig:kolmogorov-nn-out}
\end{figure}

Fig. \ref{fig:kolmogorov-full-comparison} presents a comparison of reconstruction of the vorticity $\omega$ in the Kolmogorov scenario, utilizing the various methodologies. Here, both PINNs and Kriging fail to capture the high-frequency features of the field, in contrast to SPIRE, RBF and IDW. However, IDW has again a clear point pattern, and in RBF many contours and edges are jagged rather then smooth. This is mostly corrected in the SPIRE reconstruction. Fig. \ref{fig:kolmogorov-nn-out} shows the raw NN output, before adding to the RBF and running through the solver. It shows that along such edges is where the NN amendments are most prominent.

\subsection{Runtime}
While SPIRE achieves superior reconstruction accuracy, it incurs a higher computational cost than a standalone neural network due to the integration of the differentiable solver within the training loop and during inference. Average training time per epoch and total inference time on the test set (500 samples) was compared against a U-Net with identical architecture, trained on the data without the solver-in-the-loop. For the Wake flow dataset, SPIRE required approximately 12.75 minutes per epoch compared to 5.7 minutes for the CNN. In the Kolmogorov case, 1.59 hours vs. 1.19 hours per epoch, and in Burgers, 14 seconds vs. 0.66 of a second. Inference latency was, unsurprisingly, higher when using the solver; for example, processing the Wake flow test set took 47 seconds, whereas the CNN  completed the task in roughly one second. In Kolmogorov, inference time was 19 seconds for the SPIRE model, and 3.6 seconds for the CNN baseline.

Recall that PINNs are learning the solution directly, and as such are relevant only on the trained data. Still, training time of the PINNs using separate training windows (see section \ref{sec:pinns}) was longer than training and inference on the entire test set using SPIRE. In particular, in Kolmogorov, training the 5 PINNs took 34 hours. Preprocessing of the sampled fields using RBF took slightly less than 5 hours, and only 4 additional hours for training the NN. Altogether, our pipeline is $73 \%$ faster in this scenario. In Wake, preprocessing took 20 minutes, and training duration was 1.5 hours, compared to 9 hours for the PINNs, over four times longer.

\section{Discussion}
The results presented in Table \ref{tab:results} demonstrate that SPIRE achieves superior performance on the test data compared to PINNs, IDW, and Kriging, and to RBF without application of the NN and solver. The fact that SPIRE attains the lowest error across all scenarios, consistently, underscores the strength of the approach. 

Across all case studies, SPIRE incorporates additional information from the underlying PDE into the reconstruction process, beyond merely leveraging the sample point values $s(t)$. This is emphasized by the results of the ablation, seen in Tab. \ref{tab:results}. The use of the solver alone already improved reconstruction, but in all three scenarios the trained NN further improved the results. 

First, in Fig.~\ref{fig:burgers}, SPIRE successfully maintains periodic boundary conditions, even in the absence of sensors at the domain boundaries. There, the baseline methods deviate from the true field, highlighting SPIRE’s ability to respect physical constraints even in scenarios where spatial extrapolation is desired. While PINNs do respect the BCs, the relative error and visualized example show that they are suboptimal if high reconstruction accuracy is needed. 

In Wake flow, SPIRE provides an integrated approach for handling non-trivial internal boundaries within the domain. It surpassed the compared methods in terms of relative error, with the improvement being most prominent around the cylinder obstacle itself (Fig. \ref{fig:residual-comparison}). These results suggest that SPIRE is particularly well-suited for reconstructing fields with complex geometries, which is especially advantageous for real-world applications such as modeling flow in urban environments.

Finally, in Kolmogorov flow, the NN element was shown to help both visually (Fig. \ref{fig:kolmogorov-full-comparison}) and numerically improve the reconstruction of the already next-best RBF reconstruction. It is particularly helpful in correcting sharp gradients in the vorticity field, as seen in Fig. \ref{fig:kolmogorov-nn-out}.  

Despite the apparent success of the pipeline in reconstructing the fields accurately, there are some important theoretical and practical limitations that need to be addressed.

In chaotic systems, the trajectory of the solution $u$ in the phase space is highly sensitive to changes in initial condition. The rate of divergence of similar initial conditions is often quantified using the (largest) Lyapunov exponent $\lambda\in\R$ \cite{strogatzChaos2024}. Chaotic systems are characterized by $\lambda>0$. In those cases, there is a bound on the sampling interval $\Delta t$ when applying SPIRE:
\begin{equation}
    \Delta t<\frac{1}{\lambda}
\end{equation}
Otherwise, even a close to perfect reconstruction of initial field will result in large values of loss $\sL$ (Eq. \ref{eq:final-loss}), undermining the optimization process. While exact calculation of $\lambda$ is non-trivial, the practical takeaway is that $\Delta t$ should be small compared to timescales in the given system.

Another significant limitation of the method is discretization dependence. In both training and inference, the pipeline is dependent on the choice of discretization, both spatial and temporal. The NN is trained on a particular resolution with a specific interval $\Delta t$. This distinction is most prominent when compared to the discretization independence of PINNs.
Furthermore, accuracy of the PDE solver plays a vital role. Inaccurate simulation of the initially reconstructed field may create misleading results for the NN, derailing the learning process. We speculate that the NN might be able to learn to correct errors introduced by the solver, similarly to Um et al. \cite{umSolverintheLoop2020}, but this possibility was not explored in this study.

\section{Conclusion}

In this work, we have introduced SPIRE, a framework that integrates a differentiable PDE solver within the training loop of a neural network, thereby enabling the reconstruction of complete fields from sparse measurements without requiring access to ground-truth data. Through comprehensive case studies in fluid mechanics, we demonstrate that SPIRE consistently outperforms conventional baselines such as PINNs, Kriging, IDW, and RBF methods.

Building on the present work as a proof-of-concept, future research can aim to apply and evaluate the proposed pipeline in specific engineering applications, using real-world measurement data. Moreover, generalizing the input and/or output to encompass measurements across multiple timesteps may further enhance the accuracy and broaden the applicability of the proposed methodology.

A persistent and significant challenge in applying machine learning methods to measurement data is data scarcity, particularly when contrasted with more commonly studied modalities such as text, images, and videos. The approach presented herein addresses this issue by providing a principled method to exploit spatially sparse data more effectively, tailored specifically to the complexities of the problem domain.

\section*{Software and Data}
Complete project repository is to be made available upon acceptance, including code used for generation of the simulated data.

\section*{Ethics and Privacy Statement}
The SPIRE framework offers a data‑efficient approach to reconstructing dense physical fields from sparse measurements by integrating machine learning with differentiable physics‑based simulation, potentially benefiting domains such as environmental monitoring, fluid‑dynamics‑based engineering, and scientific experimentation where high‑resolution data are costly or impractical to obtain. Its ability to enforce physical consistency may support more reliable scientific workflows and broaden access to advanced modeling in resource‑limited settings, while also serving as an educational example of physics‑guided machine learning. 

At the same time, the method relies on the correctness of the assumed governing equations and simulation models, and incorrect or overly simplified physics may lead to plausible‑looking but inaccurate reconstructions, particularly in high‑stakes or out‑of‑distribution scenarios. 

Although it raises no direct concerns regarding fairness or privacy, users should apply domain expertise and caution when interpreting outputs, recognizing that differentiable solvers introduce computational cost and that robustness under sparse sensing remains a key limitation. Overall, SPIRE contributes to the growing paradigm of combining analytical physical knowledge with modern ML, offering potential societal benefits while emphasizing the need for responsible and context‑aware use.

%%
%% The next two lines define the bibliography style to be used, and
%% the bibliography file.
\bibliographystyle{ACM-Reference-Format}
\bibliography{spire}

%%
%% If your work has an appendix, this is the place to put it.
\newpage
\appendix

\section{Additional Results}\label{sec:additional-results}
\subsection{Comparison to Neural Operators}

In recent years, there is a growing interest and research directed at development of neural operators. That is, learning a mapping between function spaces $u(t,x)\mapsto v(t,x)$, instead of learning fields $u(t,x)$ or coefficients directly as in PINNs.  As formulated in section \ref{sec:methodology}, the SPIRE pipeline can be thought of as an operator; taking as input a sparse field, and outputting a dense one. So, a comparison to well-known neural operators may be appropriate, namely DeepONets \cite{wangDeepONets2021} and Physics-Informed Neural Operators \cite{PINO}.

There are a few caveats that should be emphasized, that also prevented us from including the comparison in the results section, despite the favorable outcome for SPIRE.

DeepONets provide a general framework for a neural operator by learning mappings between infinite-dimensional function spaces. The standard architecture relies on two distinct sub-networks: a branch network that processes the input function evaluated at discrete sensor locations, and a trunk network that encodes the continuous spatial or temporal coordinates of the output domain. The final prediction is obtained by computing the dot product of the outputs from these two networks. While we include a comparison we conducted on our data for the Burgers' equation, we did not compare SPIRE with DeepONets on the Kolmogorov and Wake cases due to the lack of such examples of DeepONets in the existing literature.

PINOs were not developed for accuracy of reconstruction. Their goal is to learn the temporal evolution of the PDE, and output a spatiotemporal field $u(t,x), \: x\in D,t\in[0,T]$. As it is based on a Fourier Neural Operator (FNO) \cite{fno}, and uses a PDE residual loss, it allows for training the network on a low-resolution discretization of the domain, and then output a high-res prediction. In the particular case of choosing the sample locations $S=\{x_i\}$ on a uniform grid, this is analogous to reconstruction in some sense. Yet, we emphasize that the task of PINO is inherently different, and so as the results quoted here are taken directly from the original paper \cite{PINO} this is not a true apples-to-apples comparison. We claim that a comparison is still meaningful thanks to the nature of the relative error, that normalizes by the ground truth field. Finally, as PINO is based on a FNO, it is not trivial to apply to domains with internal boundaries \cite{luComprehensiveFairComparison2022}, such as the Wake case.

\begin{table*}[ht]
  \caption{Mean relative $L^2$ error of the reconstruction $\pm$ standard deviation, for SPIRE, PINO and DeepONets. \\
  *Results for PINO taken from \cite{PINO}.}
    \label{tab:neural-operator-results}

  \begin{center}
    \begin{small}
      \begin{sc}
        \begin{tabular}{lccccc}
        \toprule
        & \textbf{SPIRE}            & \textbf{PINO}*   & \textbf{DeepONets}                                               \\ \hline
        Burgers, $|S|=16$ & $\mathbf{0.06} \pm 0.03$   & -     & $0.09 \pm 0.03$ \\
        Burgers, $|S|=32$ & $0.017 \pm 0.006$   & $\textbf{0.002} \pm 0.0001$     & - \\
                
        Kolmogorov, $|S|=4,096$   & $\mathbf{0.04} \pm 0.01$   &      $0.06 \pm 0.001$   & -                                       \\

        \bottomrule
        \end{tabular}      
        \end{sc}
    \end{small}
  \end{center}
  \vskip -0.1in
\end{table*}

\subsection{Evaluation with Fewer Samples}
Finally, we include an evaluation of SPIRE with a smaller number of sensors ($|S|$), in Burgers' and Kolmogorov. Our claim here is that while there is no significant gain in performance, there is no loss as well, and SPIRE matches the next-best RBF reconstruction it is also based on. This underscores the robustness of the approach that uses a grey-box pipeline, where the NN $f_\theta$ is applied as a correction of the reconstruction, and not the basis for the reconstruction itself.

\begin{table*}[hb]
  \caption{Mean relative $L^2$ error of the reconstruction $\pm$ standard deviation for the $|S|=16$ Burgers and the $|S|=256$ Kolmogorov tests.}
    \label{tab:additional-results}

  \begin{center}
    \begin{small}
      \begin{sc}
        \begin{tabular}{lccccc}
        \toprule
        & \textbf{SPIRE}            & \textbf{PINN}   & \textbf{RBF}                 & \textbf{Kriging}    & \textbf{IDW}                                                \\ \hline
        Burgers, $|S|=16$ & $\mathbf{0.06} \pm 0.03$   & $0.10 \pm 0.04$     & $\mathbf{0.06} \pm 0.03$   & $0.2 \pm 0.2$ & $0.16 \pm 0.04$  \\

        Kolmogorov, $|S|=256$   & $\mathbf{0.45} \pm 0.05$   &      $1.3\pm0.2$   & $\textbf{0.45} \pm 0.05$     & $0.6 \pm 0.2$       & $0.50 \pm 0.04$                                             \\
        & &    & ($0.45 \pm 0.05$)            & ($0.6 \pm 0.2$)     & ($0.49 \pm 0.03$)                \\            

        \bottomrule
        \end{tabular}      
        \end{sc}
    \end{small}
  \end{center}
  \vskip -0.1in
\end{table*}

\begin{figure*}[hb]
\centering
\setlength{\tabcolsep}{1.5pt}
\renewcommand{\arraystretch}{1.2}

% Helper command to center images vertically on the table row baseline
\newcommand{\vcimg}[1]{%
    \raisebox{-0.5\height}{\includegraphics[width=0.16\textwidth]{#1}}%
}

\begin{tabular}{c c c c c c}
    
    % --- HEADERS ---
    \small \textbf{Ground Truth} & 
    \small \textbf{SPIRE} & \small \textbf{PINN} & \small \textbf{RBF} & \small \textbf{Kriging} & \small \textbf{IDW}  \\
    
    \cmidrule(lr){2-6} 
    
    % --- ROW 1: Images ---
    \vcimg{kolmogorov_true.png} &
    \vcimg{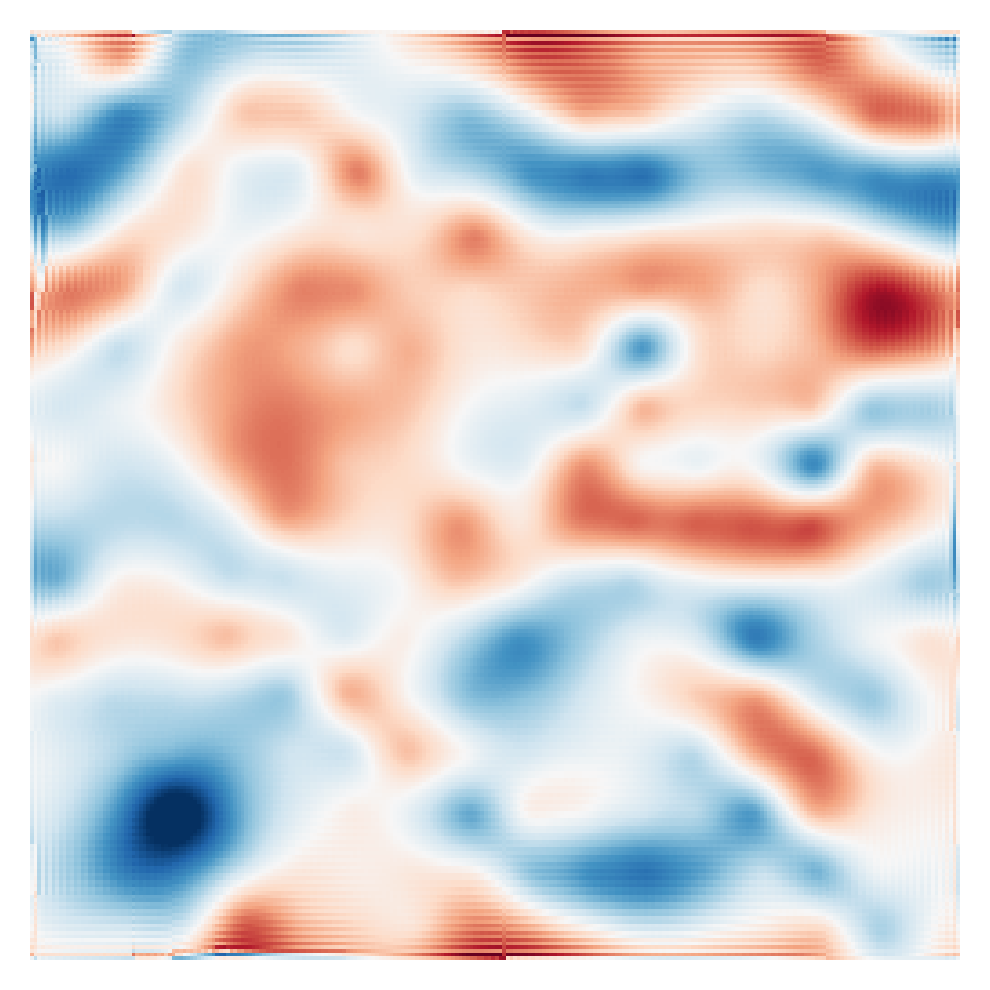} &
    \vcimg{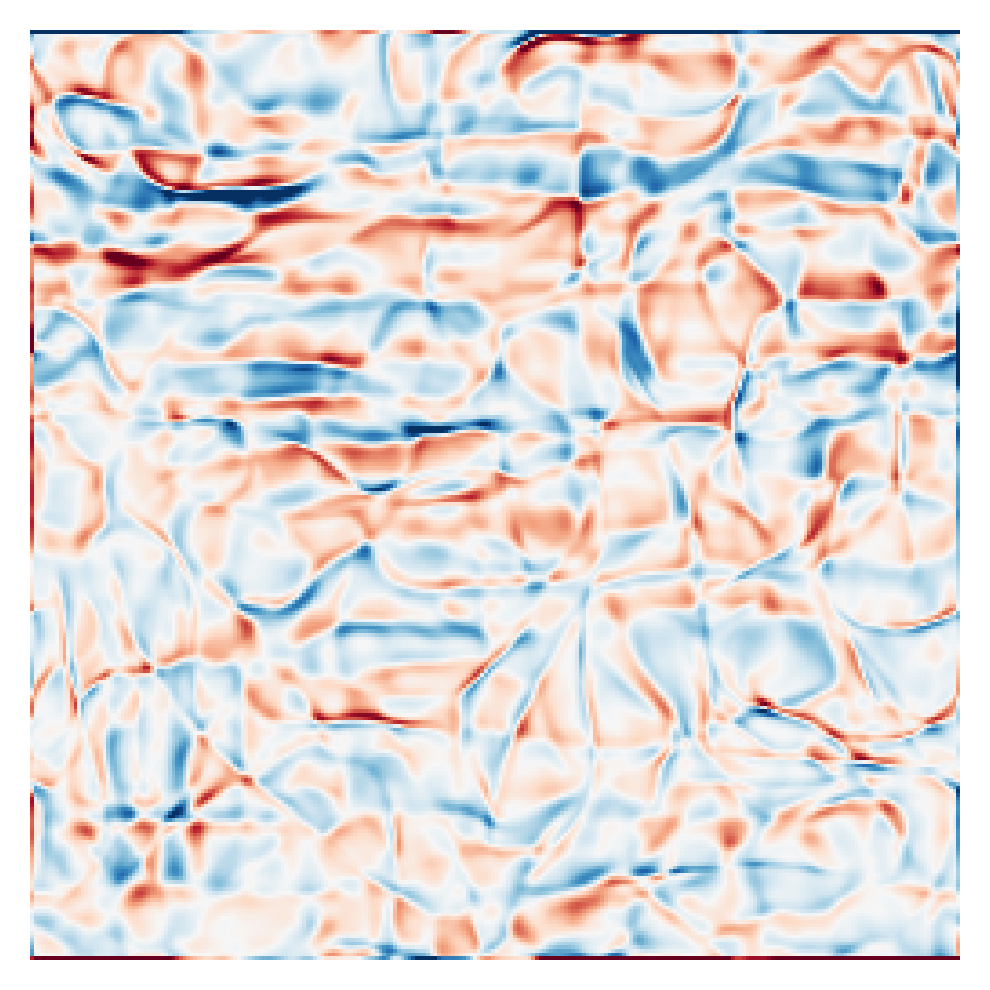} &
    \vcimg{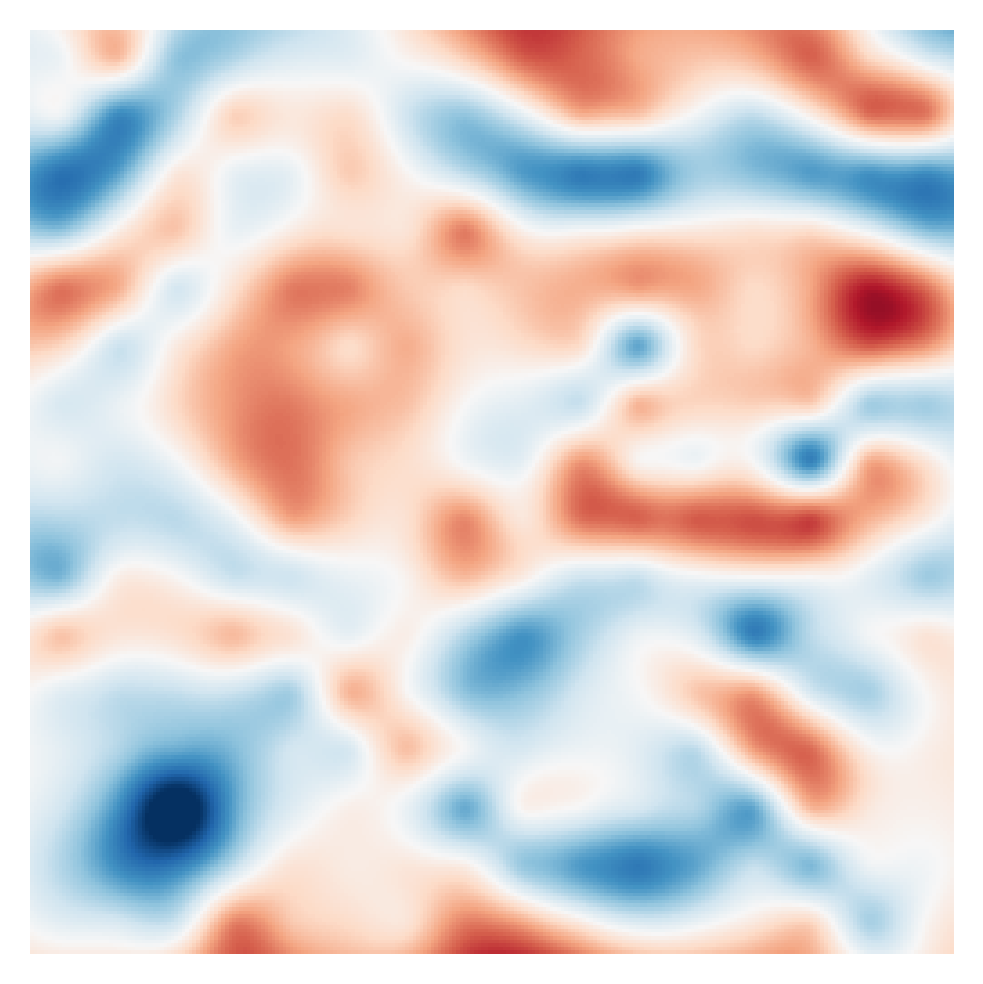} &
    \vcimg{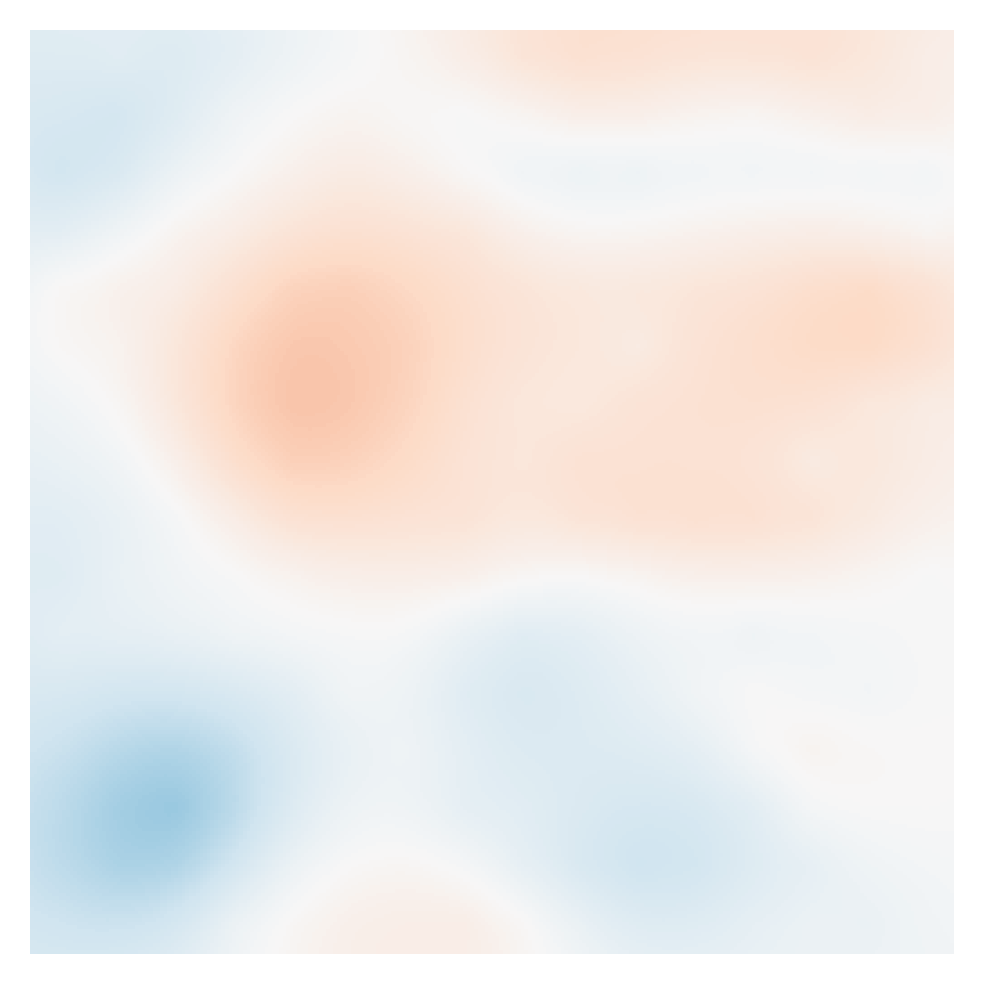} &
    \vcimg{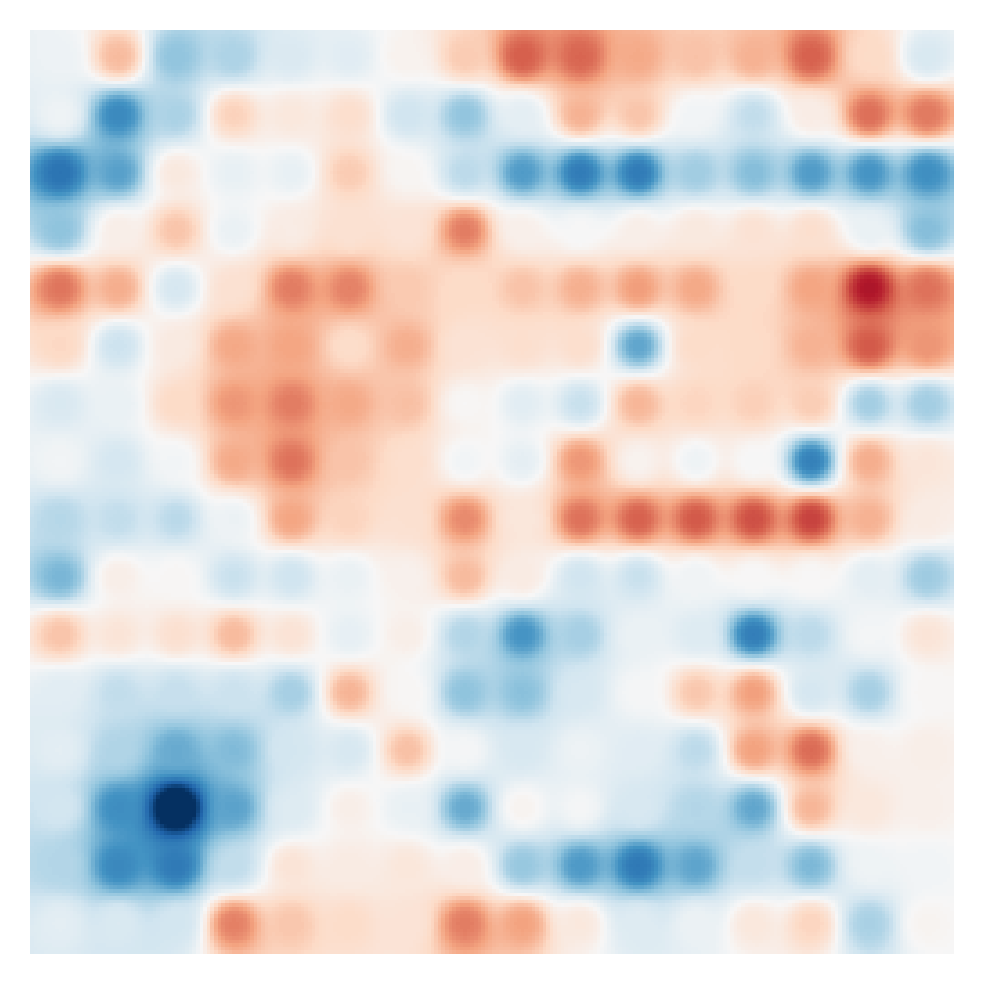}  \\
    
\end{tabular}
\Description{Plots of the reconstruction from a smaller number (256) of measurement locations, in the Kolmogorov case. RBF and SPIRE are virtually identical, while the PINN output is seemingly completely unrelated to the true field.}
\caption{Kolmogorov flow - visual comparison of reconstructions using different methods for low resolution ($|S|=256$).}
\label{fig:kolmogorov-low-res}

\end{figure*}
\end{document}